\documentclass[letterpaper,journal]{IEEEtran}

\usepackage[colorlinks,
            linkcolor=red,
            anchorcolor=blue,
            citecolor=green]{hyperref}
\usepackage{amsmath,amsfonts}
\usepackage{algorithm}
\usepackage{array}
\usepackage[caption=false,font=normalsize,labelfont=sf,textfont=sf]{subfig}
\usepackage{textcomp}
\usepackage{stfloats}
\usepackage{url}
\usepackage{verbatim}
\usepackage{graphicx}
\usepackage{cite}
\usepackage{url}
\usepackage{tabu}
\usepackage{multirow}
\usepackage{subcaption}
\usepackage{caption}
\usepackage{multicol}
\usepackage{booktabs}
\usepackage{colortbl}
\usepackage[dvipsnames,table]{xcolor}

\usepackage{algpseudocode}
\usepackage{amssymb}

\usepackage{marvosym} % \Letter

\newcommand{\method}{CLAP}
\newcommand{\bfs}[1]{\noindent \textbf{#1}}

\usepackage{amsmath,amsfonts,bm}

\def\eqref#1{equation~\ref{#1}}
\def\1{\bm{1}}

\DeclareMathAlphabet{\mathsfit}{\encodingdefault}{\sfdefault}{m}{sl}
\SetMathAlphabet{\mathsfit}{bold}{\encodingdefault}{\sfdefault}{bx}{n}

\newcommand{\add}[1]{#1}
\newenvironment{addblock}{\begingroup}{\endgroup}

\begin{document}

\title{\method: Contrastive Latent Action Pretraining for Learning Vision-Language-Action Models from Human Videos}

\author{Chubin~Zhang,
        Jianan~Wang,
        Zifeng~Gao,
        Yue~Su,
        Tianru~Dai,
        Cai~Zhou,
        Jiwen~Lu,~\IEEEmembership{Fellow,~IEEE,}
        and~Yansong~Tang,~\IEEEmembership{Member,~IEEE}%
\thanks{Chubin Zhang and Jianan Wang contributed equally. Yansong Tang is the corresponding author. Chubin Zhang, Zifeng Gao, Tianru Dai, Jiwen Lu, and Yansong Tang are with Tsinghua University. Chubin Zhang, Jianan Wang, and Yue Su are with Astribot. Yue Su is with the University of Hong Kong. Cai Zhou is with Massachusetts Institute of Technology.}%
\thanks{The code is available at \url{https://github.com/LinShan-Bin/OpenCLAP}.}}

\IEEEaftertitletext{%
\begin{center}
    \vspace{-15pt}
    \includegraphics[width=\textwidth]{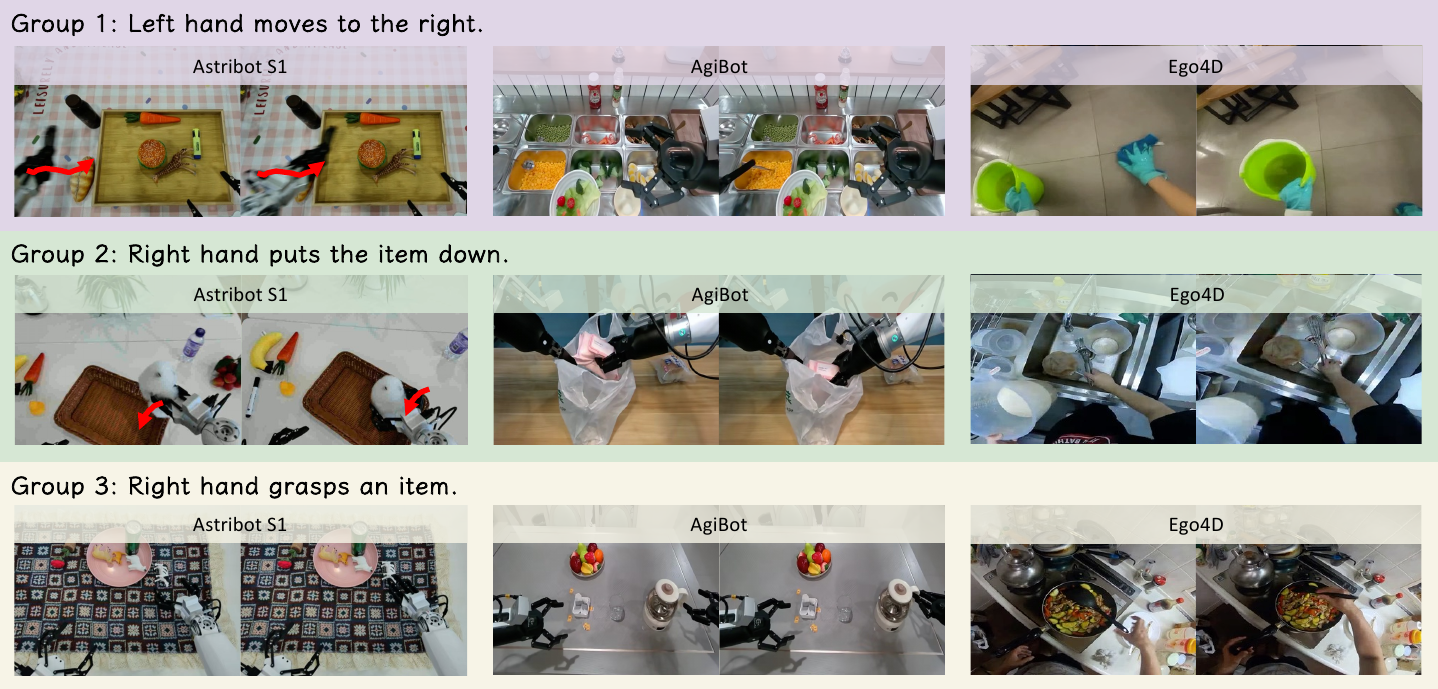}
    \captionof{figure}{\textbf{Visualization of our aligned latent action space.} We show samples from clustered action tokens, demonstrating semantic alignment across diverse robot (Astribot, AgiBot) and human (Ego4D) domains. Groups 1–3 correspond to moving right, placing, and grasping, respectively. The red arrows on the Astribot S1 frames visualize the predicted 3D trajectory decoded from the latent action and projected onto the image plane, confirming the physical executability of the learned representations.}
    \label{fig:latent_act}
    \vspace{15pt}
\end{center}
}

\maketitle

\begin{abstract}
Generalist Vision-Language-Action models remain constrained by the scarcity of robotic data relative to the abundance of human video demonstrations. Existing Latent Action Models attempt to use video data but often suffer from visual entanglement, encoding noise rather than manipulation skills. To address this limitation, we propose Contrastive Latent Action Pretraining (CLAP), a framework that first uses Act-VAE to learn an executable action-token vocabulary from robot trajectories and then aligns human visual transitions with this vocabulary through contrastive learning. This alignment maps unlabeled human videos into a physically grounded latent action space rather than reconstructing appearance. Building on the aligned tokens, we train CLAP-NTP as an autoregressive VLA using robot demonstrations and pseudo-labeled human videos, preserving instruction following and object generalization. \add{For deployment and target-domain adaptation, we further introduce a post-training strategy that combines CLAP-RF, a Rectified Flow action head for low-latency continuous action chunk prediction, with Knowledge Matching regularization to preserve pretrained semantic knowledge during fine-tuning.} Extensive experiments show that CLAP achieves strong performance against competitive baselines while enabling effective skill transfer from human videos to robotic execution.
\end{abstract}


\vspace{20pt}

\section{Introduction}
\label{sec:introduction}

\IEEEPARstart{T}{he} recent surge in Large Language Models (LLMs) and Vision-Language Models (VLMs) has demonstrated strong capabilities in semantic understanding, visual perception, and embodied reasoning~\cite{prismaticvlms,yang2024qwen2}. These advances have naturally extended to robotics, giving rise to Vision-Language-Action (VLA) models~\cite{rt-1,RT-2,openvla} as a promising approach to general-purpose manipulation. By combining the semantic knowledge of internet-scale data with embodied control, VLAs aim to create agents capable of following natural language instructions across diverse environments and tasks.

A primary obstacle in scaling VLA models is the limited availability of high-quality training data. Although large-scale robotic datasets~\cite{open_x_embodiment_rt_x_2023,bridge,rt-1,droid} have greatly benefited the community, robotic data still lag far behind human data in scale, diversity, and semantic richness. Consequently, using widely available unlabeled human videos has become a critical research direction. To address this gap, Latent Action Models (LAMs)~\cite{lapa,bu2025univla} have emerged as a popular paradigm. Existing LAMs typically use self-supervised learning to construct a latent space through inverse dynamics, predicting the latent action required to transition between adjacent video frames. While this enables learning from video, a fundamental limitation remains: these methods do not explicitly align the latent space with the robot's physical action space. As a result, the learned representation is often entangled with extraneous visual factors, such as background shifts and object deformations, rather than encoding manipulation skills. This entanglement requires complex post-hoc training to map visual latents to robot controls and severely limits direct skill transfer from human videos to robotic execution.

In this work, we address this limitation by proposing \textbf{C}ontrastive \textbf{L}atent \textbf{A}ction \textbf{P}retraining (\textbf{CLAP}). \add{Specifically, we first introduce \textbf{Act-VAE}, which quantizes continuous robot trajectories into a compact and executable action-token vocabulary. CLAP then uses contrastive learning to align visual transitions from human videos with this robot-grounded vocabulary.} Unlike prior approaches that define latent actions solely through visual reconstruction, \add{this design encourages video-derived latent actions to match the physical consequences of robot actions rather than appearance changes alone.} This alignment effectively filters out visual noise, ensuring that the latent representations extracted from human videos are isomorphic to executable robot commands.

\begin{figure}[t]
    \begin{center}
        \includegraphics[width=0.48\textwidth]{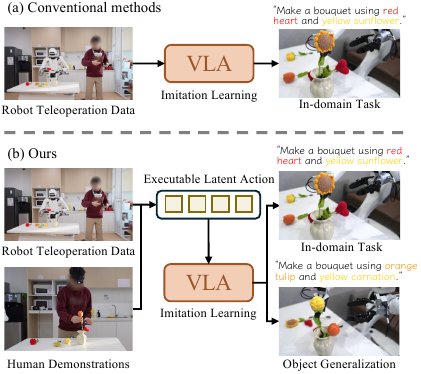}
    \end{center}
    \caption{\textbf{Overview of CLAP.} In contrast to (a) conventional methods that rely solely on limited robot teleoperation data, (b) CLAP learns an executable latent action space from large-scale human demonstrations. This enables the transfer of semantic knowledge to robot policies, achieving object generalization through human videos.}
    \label{fig:overview}
\end{figure}

\add{Building on this aligned representation, we instantiate a token-space VLA policy, \textbf{CLAP-NTP} (Next-Token Prediction). By modeling action tokens as a continuation of the language sequence, CLAP-NTP preserves the reasoning and instruction-following capabilities of the VLM backbone while learning from both robot demonstrations and pseudo-labeled human videos. This formulation is especially useful for object generalization because the policy can absorb semantic evidence from videos without requiring continuous robot actions for every new object.}

\add{For deployment and target-domain adaptation, we further introduce a unified post-training strategy that combines \textbf{CLAP-RF} (Rectified Flow~\cite{rectified_flow}) with \textbf{Knowledge Matching (KM)}. CLAP-RF attaches a flow-matching action head to the adapted NTP backbone and predicts continuous action chunks with lower policy-update latency than autoregressive token decoding. KM regularizes adaptation against a trusted NTP reference, preserving pretrained semantic knowledge while the continuous controller is tuned on robot trajectories.}

Our main contributions are summarized as follows:
\begin{itemize}
    \item \add{We introduce \textbf{Act-VAE}, which quantizes continuous robot trajectories into compact and executable action tokens, establishing a physically grounded action vocabulary.}

    \item We propose \textbf{CLAP}, a contrastive pretraining framework that aligns human visual transitions with the Act-VAE vocabulary, mitigating visual entanglement in latent action learning.

    \item We develop \textbf{CLAP-NTP}, an autoregressive VLA that learns from robot demonstrations and pseudo-labeled human videos, enabling robust instruction following and zero-shot object generalization.

    \item \add{We present an efficient post-training strategy that combines \textbf{CLAP-RF} with \textbf{Knowledge Matching}, converting token-space representations into low-latency continuous action chunks while preserving pretrained semantic knowledge during fine-tuning.}
\end{itemize}

\section{Related Work}

\subsection{Imitation Learning for Manipulation}
% Imitation learning, particularly exemplified by Behavior Cloning (BC)~\cite{jang2021bc,e2evp}, has evolved into a prevalent paradigm of robot learning, culminating in the widespread deployment of visuomotor policies~\cite{diffusionpolicy,dp3d,dsp,dspv2,action_chunk,wang2025hierarchical} for manipulation tasks. These methods typically leverage \add{maximum likelihood imitation} to model the conditional distribution from observations to actions~\cite{ddpm,ddim}, achieving remarkable success in task-specific settings. 
\add{Imitation learning, particularly Behavior Cloning (BC)~\cite{jang2021bc,e2evp}, has become a prevalent paradigm in robot learning. 
Classic BC formulates policy learning as supervised maximum likelihood estimation of actions conditioned on observations. 
Building on this objective, recent visuomotor policies~\cite{diffusionpolicy,dp3d,dsp,dspv2,action_chunk,wang2025hierarchical} often employ expressive conditional generative models, such as latent-variable models~\cite{vae} and diffusion models~\cite{ddpm,ddim}, to capture multimodal observation-to-action mappings.}
These approaches have achieved remarkable success in task-specific manipulation settings. However, the inherent heterogeneity across embodiments introduces substantial distributional diversity in the action space, which impedes broad cross-embodiment generalization~\cite{hpt}. To bridge this gap, early research sought to establish embodiment-agnostic representations such as flow~\cite{generalflow,flow2act,g3flow,atm}, object poses~\cite{mba,spot}, or atomic primitives~\cite{robogpt}, thereby decoupling the policy from specific robot kinematics. In a parallel line of work, policy-level studies have investigated retargeting strategies for transferring manipulation skills from human hands to robotic systems~\cite{maniptrans,motiontrans} or jointly learning human and robot manipulation for specific tasks~\cite{humanpolicy,dexumi}. Nevertheless, these explicit representations provide only marginal gains or depend on specialized setups, falling short of a universal solution for heterogeneous manipulation.

\subsection{Vision-Language-Action Models}
Departing from these explicit policy-level approaches, Vision-Language-Action (VLA) models~\cite{gr1,gr2,pi0,pi05,vlarl,lumo1} have shifted the field toward systematically addressing general cross-embodiment robotic manipulation~\cite{rt-1,RT-2,open_x_embodiment_rt_x_2023}. Initial VLA approaches sought to use the strong semantic priors of VLMs~\cite{prismaticvlms,yang2024qwen2} to directly fit heterogeneous action distributions~\cite{openvla,oftvla,robovlm,pi0}. However, these attempts produced suboptimal results because cross-embodiment mapping is highly complex~\cite{hpt}. In response, many studies have mitigated the issue through refined tokenization strategies~\cite{vqvla,fast,pi05,pi06} or optimized action spaces~\cite{umi,dexumi,gr3}, while others have introduced architectural enhancements such as embodiment-specific action heads~\cite{hpt,bu2025univla} and embodiment-related prompting mechanisms~\cite{xvla}. Although effective, these methods largely remain at the level of representation alignment~\cite{repa, reed}. They do not explicitly acquire primitive-level action representations and therefore struggle to distill complex behaviors into embodiment-independent quantities~\cite{mirage}.

\subsection{Latent Action Learning and Tokenization}
To address these limitations in action representation, Latent Action Models (LAMs)~\cite{lapa} have emerged as a prevailing paradigm for unifying heterogeneous action spaces. By imposing visual supervision, these methods aim to align action primitives across diverse embodiments within a shared latent manifold~\cite{semi}, which serves as an embodiment-agnostic action space~\cite{bu2025univla}. This process distills high-dimensional, multimodal actions arising from embodiment discrepancies into invariant representations that encode the underlying skills, benefiting scalable and efficient decision-making by VLMs. Technically, mainstream LAMs~\cite{lapa,bu2025univla,moto,vipra} typically employ generative~\cite{vqgan} or discriminative~\cite{dinov2,dinov3,siglip} encoders to compress action-aligned observations into a compact feature space. Through action-conditioned image reconstruction, they impose a latent structure on actions. The efficacy of this paradigm for downstream planning has been empirically validated by Agibot Go-1~\cite{go1} in large-scale training scenarios. However, current latent action models share a fundamental limitation: their latent spaces are learned from visual dynamics and are therefore susceptible to extraneous factors such as background shifts and object deformation. Consequently, the learned space is often entangled, necessitating post-hoc training for effective robotic control. This limitation prevents direct skill learning from human videos. 
\add{A complementary line of work improves action tokenization itself. BEAST~\cite{beast} uses B-spline control tokens for smooth parallel decoding, while FASTerVQ~\cite{fastervq} learns compact action-chunk tokens for efficient autoregressive VLAs. Being-H0~\cite{beingh0} further extends tokenized motion modeling to human videos through explicit hand-motion representations. These studies highlight the value of compact action vocabularies, whereas LAMs emphasize learning latent dynamics from visual transitions. Our work connects these two directions by using contrastive learning to align LAM-style human video dynamics with robot-grounded action tokens.}

\begin{figure*}[t]
    \begin{center}
        \includegraphics[width=\textwidth]{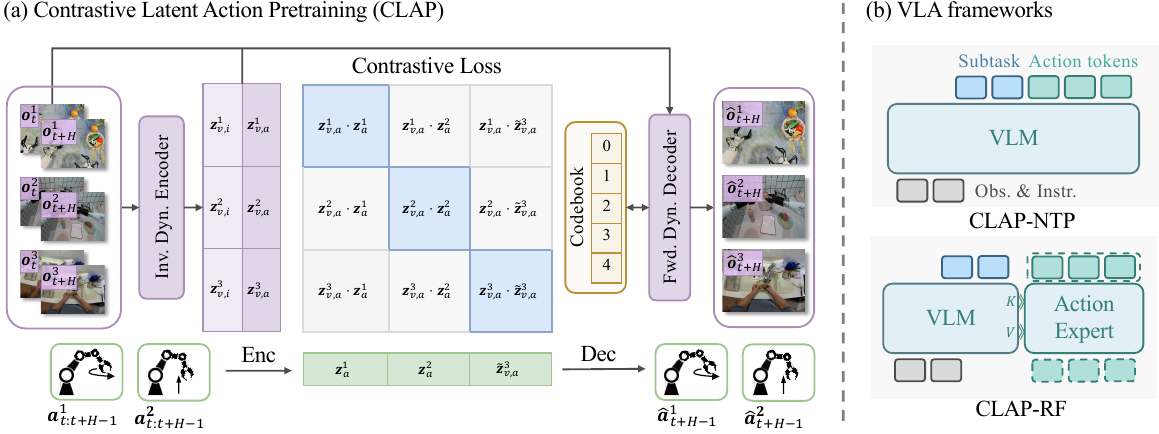}
    \end{center}
    \caption{\textbf{The pipeline of CLAP.} \textbf{(a) Contrastive Latent Action Pretraining:} Visual state transitions from videos are aligned with quantized robot actions via contrastive learning to establish a shared, physically grounded latent space. \textbf{(b) VLA Frameworks:} We introduce CLAP-NTP for discrete autoregressive planning and CLAP-RF for low-latency inference via Rectified Flow.}
    \label{fig:pipeline}
\end{figure*}

\section{Methodology}
\label{sec:method}

\subsection{Problem Formulation}
\label{sec:problem_formulation}
We address the problem of learning a generalist, language-conditioned bimanual manipulation policy by unifying large-scale human video demonstrations with precise robotic data. We consider two distinct data sources:

\begin{itemize}
    \item \textbf{Robotic Data:} Let $\mathcal{D}_{\text{rob}} = \{(\tau_i, \mathcal{I}_i)\}_{i=1}^{N_{\text{rob}}}$ denote expert robot trajectories conditioned on natural language task instructions $\mathcal{I}$. Each trajectory $\tau$ consists of a sequence of observations $\mathbf{o}_t$ and actions $\mathbf{a}_t$ over a horizon $T$.
    We focus on a dual-arm robotic setup. The action space $\mathcal{A} \subseteq \mathbb{R}^{14}$ is defined by concatenating the left ($L$) and right ($R$) arm commands. For each arm, the control input consists of the end-effector operational-space position $\mathbf{p} \in \mathbb{R}^3$, orientation (Euler angles) $\boldsymbol{\theta} \in \mathbb{R}^3$, and gripper aperture $g \in \mathbb{R}^1$. Thus, the joint action vector at time $t$ is:
    \begin{equation}
        \mathbf{a}_t = \left[ \mathbf{p}_t^L, \boldsymbol{\theta}_t^L, g_t^L, \mathbf{p}_t^R, \boldsymbol{\theta}_t^R, g_t^R \right]^\top \in \mathbb{R}^{14}.
    \end{equation}
    \item \textbf{Human Video Data:} Let $\mathcal{D}_{\text{hum}} = \{(\mathcal{V}_j, \mathcal{I}_j)\}_{j=1}^{N_{\text{hum}}}$ denote human video demonstrations. Unlike $\mathcal{D}_{\text{rob}}$, these trajectories contain only visual observations $\mathcal{V} = \{\mathbf{o}_1, \dots, \mathbf{o}_T\}$ and task annotations $\mathcal{I}$, without explicit action labels $\mathbf{a}_t$ or kinematic state information.
\end{itemize}

The core challenge is the domain gap: $\mathcal{D}_{\text{hum}}$ offers semantic diversity but lacks the kinematic grounding of $\mathcal{A}$, while $\mathcal{D}_{\text{rob}}$ provides precise dynamics but is limited in scale and diversity. Our goal is to learn a policy $\pi(\mathbf{a}_t|\mathbf{o}_t, \mathcal{I})$ that maximizes the likelihood of successful task completion by inferring a latent control manifold shared between human visual changes and robot physical actions.

\subsection{Framework Overview}
We formulate a unified Vision-Language-Action (VLA) framework that uses both the precision of robot-centric data and the semantic diversity of large-scale, unlabeled human video demonstrations. \add{Our training pipeline is structured into three stages:}
\begin{itemize}
    \item \textbf{Cross-Modal Alignment via CLAP:} We bridge the supervision gap between unlabeled human videos and labeled robot trajectories by establishing a shared latent manifold. This is achieved through \textit{Contrastive Latent Action Pretraining} (CLAP), which grounds visual state transitions from human videos in a quantized, physically executable action space. See Section~\ref{sec:method_clap} for details.
    \item \textbf{Token-Space Policy Pretraining:} \add{We train \textit{CLAP-NTP} with next-token prediction on robot action tokens and human pseudo action tokens. This allows the model to absorb both physical action labels from robots and semantic diversity from human videos. See Section~\ref{sec:method_vlas} for details.}
    \item \textbf{Post-Training for Continuous Control:} \add{After CLAP-NTP training, we optionally attach a Rectified Flow (RF) action head for low-latency inference. We further propose \textit{Knowledge Matching} (KM) fine-tuning strategy, a regularization strategy that anchors the policy update within a trusted region during the fine-tuning process. See Section~\ref{sec:method_posttrain} for details.}
\end{itemize}

\subsection{Contrastive Latent Action Pretraining (CLAP)}
\label{sec:method_clap}
A fundamental challenge in learning from heterogeneous sources is the modality mismatch: robot trajectories provide explicit actions $\mathbf{a}$, whereas human videos only exhibit visual state transitions $\mathbf{o}_t \rightarrow \mathbf{o}_{t+H}$. We propose CLAP to unify these modalities into a shared, discrete latent action space $\mathcal{Z}$, enabling the transfer of visual priors to physical control.

\subsubsection{Semantic Action Quantization (Act-VAE)}
To construct a physically grounded motion representation, we translate continuous kinematic trajectories into tokenized vocabularies. We model the action sequence $\mathbf{a}_{t:t+H-1} \in \mathbb{R}^{H \times D_a}$ using a Vector-Quantized Variational Autoencoder (VQ-VAE)~\cite{vqvae}, which we refer to as Act-VAE.

The Act-VAE consists of a Transformer-based encoder $E_\phi$ and a decoder $D_\psi$. \add{The encoder first embeds the action trajectory into action features, concatenates them with $N_q$ learnable register tokens, and applies self-attention over the joint sequence. The updated register tokens form $N_q$ continuous latents $\mathbf{z}_e=[\mathbf{z}_{e,1},\ldots,\mathbf{z}_{e,N_q}]$.} These latents are discretized through a learnable action codebook $\mathcal{C}_{\text{act}} = \{\mathbf{e}_k\}_{k=1}^{K}$. Each latent vector is replaced by its nearest codebook neighbor, yielding a quantized latent sequence $\mathbf{z}_q$ and the corresponding discrete token sequence $\mathbf{z}_a$. The objective minimizes the reconstruction error together with the codebook and commitment losses:
\begin{equation}
\begin{aligned}
    \mathcal{L}_{\text{Act}} = {}& \| \mathbf{a} - D_\psi(\mathbf{z}_q) \|_2^2 + \| \operatorname{sg}(E_\phi(\mathbf{a})) - \mathbf{z}_q \|_2^2 \\
    & + \beta \| E_\phi(\mathbf{a}) - \operatorname{sg}(\mathbf{z}_q) \|_2^2,
\end{aligned}
\label{eq:act_vae_loss}
\end{equation}
where $\operatorname{sg}(\cdot)$ denotes the stop-gradient operator. By optimizing the codebook size $K$ and sequence length $N_q$, we achieve a representation that balances semantic compactness with the granularity required for precise manipulation, effectively creating a ``physical language'' for the VLM and a latent space for further alignment. The frozen Act-VAE encoder used below is denoted as $E_{\text{act}}$.

% Algorithm 1: Action VQ-VAE Training
\begin{algorithm}[t]
\caption{Action VQ-VAE (Act-VAE) Training}
\label{alg:act_vae}
\begin{algorithmic}[1]
\Require Dataset of action trajectories $\mathcal{D}_{\text{act}}$, Codebook size $K$, Commitment $\beta$
\State Initialize encoder $E_\phi$, decoder $D_\psi$, action codebook $\mathcal{C}_{\text{act}} = \{\mathbf{e}_k\}_{k=1}^{K}$
\While{not converged}
    \State Sample action batch $\mathbf{a}_{t:t+H-1} \sim \mathcal{D}_{\text{act}}$
    \State $\mathbf{z}_e \gets E_\phi(\mathbf{a}_{t:t+H-1})$ \Comment{Encode to continuous latents}
    \State $\mathbf{z}_q \gets Q(\mathbf{z}_e;\mathcal{C}_{\text{act}})$ \Comment{Nearest-neighbor lookup}
    \State $\hat{\mathbf{a}}_{t:t+H-1} \gets D_\psi(\mathbf{z}_q)$ \Comment{Reconstruct trajectory}
    
    \State \textbf{Compute Loss:}
    \State $\mathcal{L}_{\text{rec}} \gets \| \mathbf{a}_{t:t+H-1} - \hat{\mathbf{a}}_{t:t+H-1} \|_2^2$
    \State $\mathcal{L}_{\text{code}} \gets \| \operatorname{sg}(\mathbf{z}_e) - \mathbf{z}_q \|_2^2 + \beta \| \mathbf{z}_e - \operatorname{sg}(\mathbf{z}_q) \|_2^2$
    \State $\mathcal{L}_{\text{Act}} \gets \mathcal{L}_{\text{rec}} + \mathcal{L}_{\text{code}}$
    
    \State Update $\phi, \psi, \mathcal{C}_{\text{act}}$ via gradient descent on $\mathcal{L}_{\text{Act}}$
\EndWhile
\end{algorithmic}
\end{algorithm}

\begin{algorithm}[t]
\caption{Vision-Dynamic VQ-VAE (VD-VAE) Training}
\label{alg:vd_vae}
\begin{algorithmic}[1]
\Require Robot data $\mathcal{D}_{\text{rob}}=\{(\mathbf{o}_t,\mathbf{o}_{t+H},\mathbf{a}_{\text{gt}})\}$,
human video data $\mathcal{D}_{\text{hum}}=\{(\mathbf{o}_t,\mathbf{o}_{t+H})\}$,
frozen visual backbone $V$,
frozen Act-VAE encoder $E_{\text{act}}$ and codebook $\mathcal{C}_{\text{act}}$,
learnable environment codebook $\mathcal{C}_{\text{env}}$,
EMA rate $\mu$
\State Initialize inverse encoder $E_{\text{inv}}$ and forward decoder $D_{\text{fwd}}$
\State Initialize teacher encoder $E_{\text{inv}}^{\text{tea}} \gets E_{\text{inv}}$
\While{not converged}
    \State Sample a mixed mini-batch $\mathcal{B}$ from $\mathcal{D}_{\text{rob}} \cup \mathcal{D}_{\text{hum}}$
    \ForAll{$s \in \mathcal{B}$}
        \State Extract transition $(\mathbf{o}_t,\mathbf{o}_{t+H})$ and source flag
        \State $\mathbf{f}_t,\mathbf{f}_{t+H} \gets V(\mathbf{o}_t),V(\mathbf{o}_{t+H})$
        \State \add{$\mathbf{o}'_t,\mathbf{o}'_{t+H} \gets \text{Augment}(\mathbf{o}_t,\mathbf{o}_{t+H})$}
        \State \add{$\mathbf{f}'_t,\mathbf{f}'_{t+H} \gets V(\mathbf{o}'_t),V(\mathbf{o}'_{t+H})$}
        \State $\mathbf{z}_{v,a},\mathbf{z}_{v,i} \gets E_{\text{inv}}(\mathbf{f}'_t,\mathbf{f}'_{t+H})$
        \State $\mathbf{z}_{q,a} \gets Q(\mathbf{z}_{v,a};\mathcal{C}_{\text{act}})$
        \State $\mathbf{z}_{q,i} \gets Q(\mathbf{z}_{v,i};\mathcal{C}_{\text{env}})$
        \State $\hat{\mathbf{f}}_{t+H} \gets D_{\text{fwd}}(\mathbf{f}_t,\mathbf{z}_{q,a},\mathbf{z}_{q,i})$

        \State $\mathcal{L}_{\text{rec}} \gets \|\mathbf{f}_{t+H}-\hat{\mathbf{f}}_{t+H}\|_2^2$
        \State $\mathcal{L}_{\text{VQ}} \gets \operatorname{VQ}(\mathbf{z}_{v,a};\mathcal{C}_{\text{act}})+\operatorname{VQ}(\mathbf{z}_{v,i};\mathcal{C}_{\text{env}})$
        \State $\mathcal{L}_{\text{reg}} \gets \|\mathbf{z}_{v,i}\|_1$

        \If{\add{$s \in \mathcal{D}_{\text{rob}}$}}
            \State \add{Read $\mathbf{a}_{\text{gt}}$ from $s$ and compute $\mathbf{z}_{a}^{+} \gets \operatorname{sg}(E_{\text{act}}(\mathbf{a}_{\text{gt}}))$}
        \Else
            \State \add{$(\tilde{\mathbf{z}}_{v,a},\tilde{\mathbf{z}}_{v,i}) \gets E_{\text{inv}}^{\text{tea}}(\mathbf{f}_t,\mathbf{f}_{t+H})$}
            \State \add{$\mathbf{z}_{a}^{+} \gets \operatorname{sg}(\tilde{\mathbf{z}}_{v,a})$}
        \EndIf
        \State $\mathcal{L}_{\text{con}}(s) \gets \operatorname{SigLIP}(\mathbf{z}_{v,a},\mathbf{z}_{a}^{+})$
        
        \State $\mathcal{L}(s) \gets \mathcal{L}_{\text{rec}} + \lambda_{\text{vq}}\mathcal{L}_{\text{VQ}} + \lambda_{\text{con}}\mathcal{L}_{\text{con}}(s) + \lambda_{\text{reg}}\mathcal{L}_{\text{reg}}$
    \EndFor

    \State $\mathcal{L}_{\text{batch}} \gets \frac{1}{|\mathcal{B}|}\sum_{s\in\mathcal{B}}\mathcal{L}(s)$
    \State Update $E_{\text{inv}},D_{\text{fwd}},\mathcal{C}_{\text{env}}$ by gradient descent on $\mathcal{L}_{\text{batch}}$
    \State Update $E_{\text{inv}}^{\text{tea}} \gets \mu E_{\text{inv}}^{\text{tea}}+(1-\mu)E_{\text{inv}}$ with EMA
\EndWhile
\end{algorithmic}
\end{algorithm}

\subsubsection{Cross-Modal Dynamics Alignment (VD-VAE)}
To harness unlabeled video data, we introduce the Vision-Dynamic VQ-VAE (VD-VAE), which infers latent actions solely from visual evolution. The VD-VAE functions as an inverse dynamics model, mapping the transition between frames $\mathbf{o}_t$ and $\mathbf{o}_{t+H}$ to the pre-established action codebook $\mathcal{C}_{\text{act}}$. \add{In this module, robot transitions are written as $\mathcal{D}_{\text{rob}}=\{(\mathbf{o}_t,\mathbf{o}_{t+H},\mathbf{a}_{\text{gt}})\}$, while human video transitions are written as $\mathcal{D}_{\text{hum}}=\{(\mathbf{o}_t,\mathbf{o}_{t+H})\}$.}

Let $\mathbf{f}_t, \mathbf{f}_{t+H}$ be visual features extracted by a frozen backbone (e.g., DINO\cite{dinov3}). An inverse dynamics encoder decomposes the transition into two disentangled latent streams: an \textit{action-relevant} latent sequence $\mathbf{z}_{v,a}$ and an \textit{action-irrelevant} latent sequence $\mathbf{z}_{v,i}$. Crucially, we enforce that $\mathbf{z}_{v,a}$ aligns with the robot's control space by quantizing it using the \textit{frozen} Act-VAE codebook $\mathcal{C}_{\text{act}}$. Conversely, $\mathbf{z}_{v,i}$ captures nuisance variables (e.g., background changes) using a separate learnable environment codebook $\mathcal{C}_{\text{env}}$. \add{A forward decoder reconstructs the clean future feature from the current feature and the quantized latents, i.e., $\hat{\mathbf{f}}_{t+H}=D_{\text{fwd}}(\mathbf{f}_t,\mathbf{z}_{q,a},\mathbf{z}_{q,i})$, providing dense supervision in addition to contrastive alignment.}

To semantically ground the visual latent to physical actions, we employ a contrastive loss to align the continuous vision-based latent sequence $\mathbf{z}_{v,a}$ with a source-specific reference action latent sequence $\mathbf{z}_{a}^{+}$. \add{For robot samples, $\mathbf{z}_{a}^{+}=\operatorname{sg}(E_{\text{act}}(\mathbf{a}_{\text{gt}}))$ is the stop-gradient Act-VAE encoding of the ground-truth action chunk. For human videos, where no robot action label is available, we use an EMA teacher encoder on the clean transition to construct a stop-gradient pseudo-positive, $\mathbf{z}_{a}^{+}=\operatorname{sg}(E_{\text{inv}}^{\text{tea}}(\mathbf{f}_t,\mathbf{f}_{t+H})_a)$, while the online branch receives augmented features. This avoids contrasting $\mathbf{z}_{v,a}$ with itself from the same online pass.} Following CLIP-style contrastive representation learning~\cite{radford2021learning}, we utilize the Sigmoid Loss for Language-Image Pre-training, or SigLIP~\cite{siglip}, which optimizes pairwise binary classification. For a positive pair $(\mathbf{z}_{v,a}, \mathbf{z}_{a}^{+})$ and a set of $M$ negative reference latents $\{\mathbf{z}_{a,j}^{-}\}_{j=1}^M$ from other samples in the batch, the loss is defined as:
\begin{equation}
\begin{aligned}
    \mathcal{L}_{\text{contrastive}} =& - \log \sigma \left( \frac{s_p - b}{\tau_{\text{c}}} \right) \\
    &- \sum_{j=1}^{M} \log \left(1 - \sigma \left( \frac{s_{n,j} - b}{\tau_{\text{c}}} \right) \right),
\end{aligned}
\end{equation}
where $s_p = \operatorname{sim}(\mathbf{z}_{v,a}, \mathbf{z}_{a}^{+})$ and $s_{n,j} = \operatorname{sim}(\mathbf{z}_{v,a}, \mathbf{z}_{a,j}^{-})$ are cosine similarities, $\tau_{\text{c}}$ is the contrastive temperature, and $b$ is a learnable bias.

Moreover, to enforce the desired disentanglement and avoid unnecessary usage of action-irrelevant latents, we apply L1 regularization to the action-irrelevant latent sequence, $\mathcal{L}_{\text{reg}} = \|\mathbf{z}_{v,i}\|_1$, encouraging sparsity and forcing it to capture only nuisance information and leave most action-relevant information in $\mathbf{z}_{v,a}$. The total objective combines dynamics reconstruction, VQ constraints, contrastive alignment and L1 regularization of the action-irrelevant latent:
\begin{equation}
\begin{aligned}
    \mathcal{L}_{\text{VD}} =& \mathcal{L}_{\text{rec}}(\hat{\mathbf{f}}_{t+H}) + \lambda_{\text{vq}} \mathcal{L}_{\text{VQ}} \\
    &+ \lambda_{\text{con}} \mathcal{L}_{\text{contrastive}} + \lambda_{\text{reg}} \| \mathbf{z}_{v,i} \|_1,
\end{aligned}
\end{equation}
where $\mathcal{L}_{\text{rec}}=\|\mathbf{f}_{t+H}-\hat{\mathbf{f}}_{t+H}\|_2^2$, and $\lambda_{\text{reg}}$, $\lambda_{\text{vq}}$, and $\lambda_{\text{con}}$ are hyperparameters weighting the regularization, VQ, and contrastive terms, respectively.
\add{After VD-VAE training, we denote the discrete pseudo action tokens inferred from human video transitions as $\hat{\mathbf{z}}_a$.}

% Algorithm 3: CLAP-NTP Training
\begin{algorithm}[t]
\caption{CLAP-NTP Training}
\label{alg:clap_ar}
\begin{algorithmic}[1]
\Require Robot Data $\mathcal{D}_{\text{rob}}$, Human Videos $\mathcal{D}_{\text{hum}}$, Trained VD-VAE
\State Initialize NTP policy $\phi_{\theta}$
\While{not converged}
    \State Sample $s=(\mathcal{I}, \mathbf{o}_t, \text{trajectory}) \sim \mathcal{D}_{\text{rob}} \cup \mathcal{D}_{\text{hum}}$
    
    \If{$s \in \mathcal{D}_{\text{rob}}$}
        \State $Y \gets [\text{subtask}, \mathbf{z}_a(\text{trajectory})]$
    \Else \Comment{Source is Human Video}
        \State $Y \gets [\text{subtask}, \hat{\mathbf{z}}_a(\text{trajectory})]$
    \EndIf
    
    \State Predict logits $\hat{y}_{\ell} = \phi_{\theta}(y_{<\ell}, \mathbf{o}_t, \mathcal{I})$
    \State $\mathcal{L}_{\text{NTP}} \gets - \sum_{\ell=1}^{L_y} \log P_\theta(y_{\ell} | y_{<\ell}, \mathbf{o}_t, \mathcal{I})$
    \State Update $\theta$ to minimize $\mathcal{L}_{\text{NTP}}$
\EndWhile
\end{algorithmic}
\end{algorithm}

\begin{addblock}
\subsection{CLAP-NTP: Token-Space Policy Learning}
\label{sec:method_vlas}
\add{Building on the aligned latent space, we first train CLAP-NTP as a token-space generalist policy.}
\end{addblock}

CLAP-NTP exploits the reasoning capabilities of VLMs to decompose complex instructions $\mathcal{I}$ into intermediate sub-goals and discrete action tokens. \add{Modeled as an autoregressive generator with parameters $\theta$, it predicts the joint sequence of sub-tasks and action indices $Y = [\mathbf{y}_{\text{sub}}, \mathbf{z}_a]$ from current observations.}
We train CLAP-NTP via next-token prediction:
\begin{equation}
    \add{\mathcal{L}_{\text{NTP}} = - \sum_{\ell=1}^{L_y} \log P_\theta(y_\ell | y_{<\ell}, \mathbf{o}_t, \mathcal{I}).}
\end{equation}
\add{Here $\ell$ indexes the autoregressive token position in the VLM output sequence of length $L_y$.}
\add{This stage unifies robot demonstrations (using ground-truth $\mathbf{z}_a$) and human videos (using VD-VAE pseudo action tokens $\hat{\mathbf{z}}_a$) for training.} Since the NTP model shares the training paradigm of the base VLM, it preserves the model's reasoning faculties, enabling direct robot control with robust instruction following. \add{After this stage, the trained NTP policy is denoted as $\phi_{\text{pre}}$ for post-training.}

\begin{figure}[t]
    \begin{center}
        \includegraphics[width=0.48\textwidth]{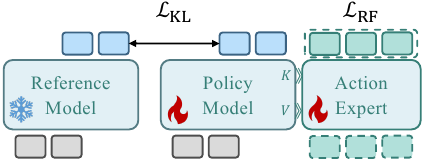}
    \end{center}
    \caption{\textbf{Knowledge Matching algorithm.} \textcolor{gray}{Gray blocks} represent the input observations and instructions. \textcolor{cyan}{Blue blocks} denote the subtask and discrete action tokens, where $\mathcal{L}_{\text{KL}}$ constrains the policy distribution. \textcolor{teal}{Green blocks} represent the continuous actions, trained via $\mathcal{L}_{\text{RF}}$.}
    \label{fig:method_km}
\end{figure}

\begin{addblock}
\subsection{Efficient Post-Training Strategy}
\label{sec:method_posttrain}
\add{After CLAP-NTP has been trained in token space, we use two complementary post-training components for target-domain adaptation.}
\add{CLAP-RF denotes the Rectified Flow action head that converts NTP representations into low-latency continuous action chunks, while Knowledge Matching (KM) regularizes the trainable token policy against a frozen NTP reference.}
\add{We first define these two components and then summarize the post-training procedure in Algorithm~\ref{alg:km_ft}.}
\end{addblock}

\subsubsection{CLAP-RF: Flow Matching Head for Fast Inference}
Autoregressive token decoding is inherently slow for deployment. CLAP-RF takes the CLAP-NTP as an understanding backbone and attaches a Diffusion Transformer (DiT)-based RF action head~\cite{dit}, denoted $f^a$, to predict continuous action chunks. The RF action head queries the NTP backbone's internal representations by attending to its key-value cache via cross-attention. The RF action head is trained by minimizing a rectified flow loss on robot trajectories. For a given robot action chunk $\mathbf{a}_{t:t+H-1}$, we first create a noised version $\mathbf{a}^{\rho}_{t:t+H-1} = \rho \mathbf{a}_{t:t+H-1} + (1 - \rho)\boldsymbol{\epsilon}$, where $\boldsymbol{\epsilon} \sim \mathcal{N}(0, I)$ and $\rho \in [0, 1]$ is the flow time. The model is trained to predict the vector field $\mathbf{v} = \mathbf{a}_{t:t+H-1} - \boldsymbol{\epsilon}$. The loss function is defined as:
\begin{equation}
    \mathcal{L}_{\text{RF}} = \mathbb{E}_{\mathcal{D}_{\text{rob}}, \rho, \boldsymbol{\epsilon}} \left[ \left\| (\mathbf{a}_{t:t+H-1} - \boldsymbol{\epsilon}) - f^a(\mathbf{a}^{\rho}_{t:t+H-1}, \rho, \mathbf{h}_t) \right\|^2 \right]
\end{equation}
where $\mathbf{h}_t$ is the RF conditioning representation obtained from the NTP backbone. Thus, CLAP-RF receives continuous supervision from robot trajectories and benefits from human-video pretraining through the NTP representations that condition the RF head.

\subsubsection{Knowledge Matching: Regularized Adaptation}
\label{sec:method_km}
Fine-tuning pretrained policies on specific embodiments often leads to catastrophic forgetting of the pretrained priors. We address this via \textit{Knowledge Matching} (KM), \add{a regularization strategy that anchors the token-policy update within a trusted region.}

\add{KM operates on the autoregressive token policy that conditions the RF action head.}
\add{Let $\phi_{\text{ref}}$ denote a frozen reference NTP policy and $\phi_{\text{policy}}$ denote the trainable token-policy branch used during RF post-training.}
\add{The token-level conditioning context is defined as $c_\ell=(y_{<\ell},\mathbf{o}_t,\mathcal{I})$.}
\add{KM is applied to autoregressive distributions over discrete tokens using a reverse KL term:}
\begin{equation}
\begin{aligned}
\mathcal{L}_{\text{KL}}
&= \sum_{\ell=1}^{L_y} D_{\text{KL}}
\!\left(
p_{\phi_{\text{policy}}}(y_\ell \mid c_\ell)
\,\|\,
p_{\phi_{\text{ref}}}(y_\ell \mid c_\ell)
\right),\\
\mathcal{L}_{\text{post}}
&= \mathcal{L}_{\text{RF}}+\alpha\mathcal{L}_{\text{KL}}.
\end{aligned}
\end{equation}
where $y_\ell$ is a discrete token at autoregressive position $\ell$ and $\mathcal{L}_{\text{RF}}$ trains the CLAP-RF head on continuous robot action chunks.

\add{This regularization allows the RF controller to adapt to the target embodiment while keeping the trainable token policy close to the trusted NTP reference.}

\begin{addblock}
\subsubsection{Post-Training Procedure}
\add{Algorithm~\ref{alg:km_ft} summarizes the complete target-domain procedure.}
\add{Starting from the pretrained CLAP-NTP policy $\phi_{\text{pre}}$, Stage~1 optionally adapts the NTP policy on all tokenized target data, including robot action tokens from $\mathcal{D}_{\text{rob}}^{\text{new}}$ and human pseudo action tokens from $\mathcal{D}_{\text{hum}}^{\text{new}}$.}
\add{The adapted NTP policy is then frozen as $\phi_{\text{ref}}$.}
\add{In Stage~2, we initialize the trainable policy branch $\phi_{\text{policy}}$ and the RF action head from this adapted model, and train them only on robot trajectories because continuous action chunks are available only in $\mathcal{D}_{\text{rob}}^{\text{new}}$.}
\add{Human videos influence CLAP-RF through the token-space NTP adaptation and the KM reference used during robot-only RF post-training.}
\end{addblock}

% Algorithm 4: Post-training with RF and Knowledge Matching
\begin{algorithm}[t]
\caption{\add{Post-training with RF and Knowledge Matching}}
\label{alg:km_ft}
\begin{algorithmic}[1]
\Require \add{Pretrained NTP policy $\phi_{\text{pre}}$, target robot data $\mathcal{D}_{\text{rob}}^{\text{new}}$, target human videos $\mathcal{D}_{\text{hum}}^{\text{new}}$, trained Act-VAE and VD-VAE, KM weight $\alpha$}
\State Initialize $\phi_{\text{NTP}} \gets \phi_{\text{pre}}$
\Statex \textbf{Stage 1: all-data NTP adaptation}
\While{NTP adaptation not converged}
    \State Sample $s \sim \mathcal{D}_{\text{rob}}^{\text{new}} \cup \mathcal{D}_{\text{hum}}^{\text{new}}$
    \If{$s \in \mathcal{D}_{\text{rob}}^{\text{new}}$}
        \State Build target tokens $Y$ from Act-VAE robot action tokens
    \Else
        \State Build target tokens $Y$ from VD-VAE pseudo action tokens
    \EndIf
    \State Update $\phi_{\text{NTP}}$ by minimizing $\mathcal{L}_{\text{NTP}}(Y)$
\EndWhile
\State Set frozen reference $\phi_{\text{ref}} \gets \phi_{\text{NTP}}$
\State \add{Initialize RF action head $f^a$ and trainable policy branch $\phi_{\text{policy}} \gets \phi_{\text{NTP}}$}
\Statex \add{\textbf{Stage 2: robot-only RF post-training with KM}}
\While{RF post-training not converged}
    \State Sample $(\mathcal{I},\mathbf{o}_t,\mathbf{a}_{t:t+H-1}) \sim \mathcal{D}_{\text{rob}}^{\text{new}}$
    \State \add{Compute RF conditioning representation $\mathbf{h}_t$ from $\phi_{\text{policy}}(\mathbf{o}_t,\mathcal{I})$}
    \State \add{Compute $\mathcal{L}_{\text{RF}}$ on continuous robot chunk $\mathbf{a}_{t:t+H-1}$}
    \State \add{Build robot action-token context $c_\ell=(y_{<\ell},\mathbf{o}_t,\mathcal{I})$}
    \State \add{Compute reverse KL regularization $\mathcal{L}_{\text{KL}}$}
    \State \add{$\mathcal{L}_{\text{post}} \gets \mathcal{L}_{\text{RF}}+\alpha\mathcal{L}_{\text{KL}}$}
    \State \add{Update $f^a$ and $\phi_{\text{policy}}$ to minimize $\mathcal{L}_{\text{post}}$}
\EndWhile
\end{algorithmic}
\end{algorithm}

\begin{figure}[t]
    \begin{center}
        \includegraphics[width=0.48\textwidth]{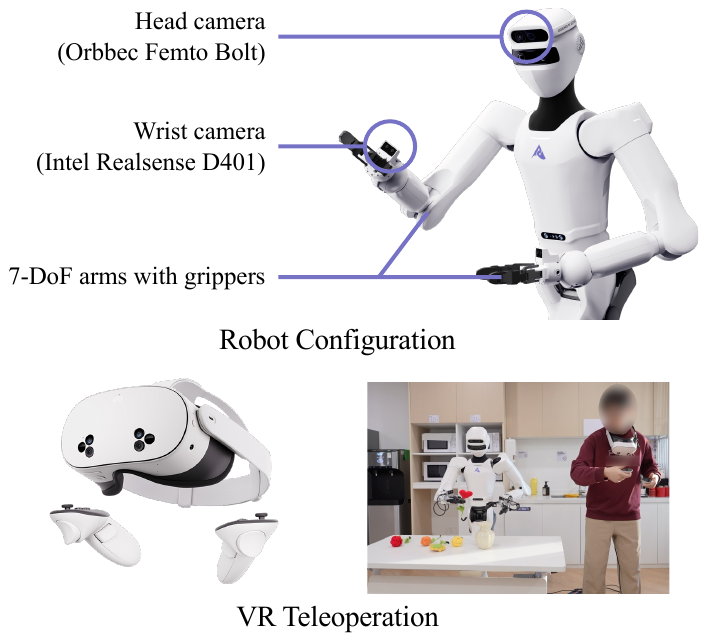}
    \end{center}
    \caption{\textbf{Experimental setup.} The \textbf{Robot Configuration} (top) features the Astribot S1 with dual 7-DoF arms and a multi-camera perception suite. \textbf{VR Teleoperation} (bottom) is performed using a Meta Quest 3S headset to collect human demonstration data.}
    \label{fig:exp_setup}
\end{figure}

\section{Model Training Details}
\label{sec:exp_pretrain}

In this section, we describe the dataset construction and training schedule for the proposed CLAP framework. Detailed model parameters are provided in Table~\ref{tab:params}.

\subsection{Dataset}
\label{sec:model_dataset}
To align with our objective of learning generalist manipulation policies from heterogeneous sources, we pretrain our latent action model using a combination of labeled bimanual robotic data and unlabeled human video demonstrations. The composite dataset comprises the following sources:

\begin{enumerate}
    \item \textbf{Curated AgiBot World Beta~\cite{go1}:} This large-scale robotic manipulation dataset contains approximately 1 million trajectories ($\sim$ 3,000 hours) spanning 217 tasks and 106 scenes (e.g., domestic, industrial, and retail environments). Data were collected using AgiBot G1 dual-arm humanoids equipped with 7-DoF arms and dexterous end-effectors. For our experiments, we use a curated subset to ensure high-quality supervision. We filter out mobile manipulation, cooperative tasks, dexterous hand data, and tasks with semantic ambiguity. The resulting subset comprises approximately 100,000 episodes, totaling 1,500 hours of high-quality bimanual interaction data.
    
    \item \textbf{Self-collected Astribot S1 Data:} To facilitate cross-embodiment adaptation, we introduce a dataset collected on the Astribot S1 platform~\cite{gao2025towards}. The robot features two 7-DoF arms with parallel-jaw grippers and a perception suite consisting of an Orbbec Femto Bolt head camera, an Orbbec Gemini 335 torso camera, and wrist-mounted Intel Realsense D401 cameras. Expert demonstrations were acquired via VR teleoperation (Meta Quest 3S), with the head camera actively tracking the workspace center. We focus primarily on pick-and-place tasks involving 90 distinct objects. This dataset contains 27,000 episodes, amounting to approximately 50 hours of data recorded at 30 Hz.
    
    \item \add{\textbf{DROID~\cite{droid}:} DROID is a large-scale in-the-wild robot manipulation dataset containing approximately 76,000 demonstration trajectories, totaling 350 hours of video data collected across diverse real-world environments. We incorporate DROID to improve single-arm manipulation performance with additional robot supervision. To fit our unified dual-arm action format, we treat DROID trajectories as right-arm demonstrations and zero-pad the left-arm action channels.}
    
    \item \textbf{Ego4D~\cite{ego4d} Human Data:} To use large-scale human priors, we employ Ego4D, a massive egocentric video dataset covering diverse daily activities. Specifically, we use the subset provided by UniVLA~\cite{bu2025univla}, which consists of 90 hours of curated trajectories relevant to manipulation tasks.
\end{enumerate}

\begin{figure}[t]
    \begin{center}
        \includegraphics[width=0.4\textwidth]{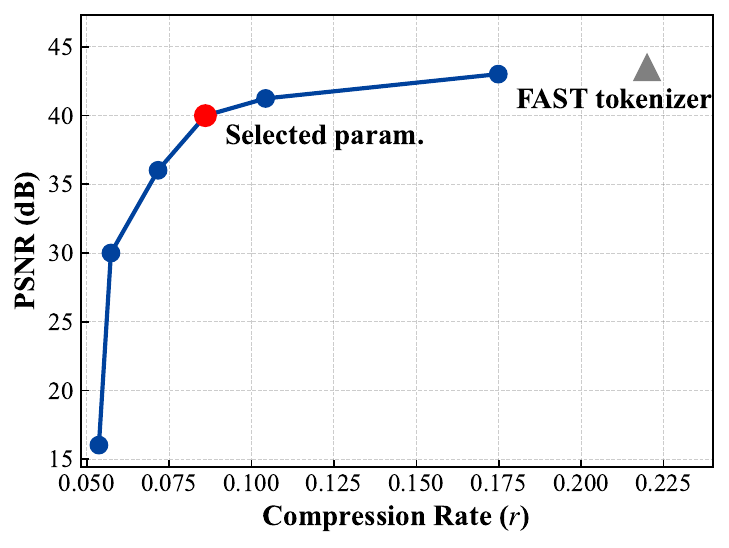}
    \end{center}
    \caption{\textbf{Rate-distortion analysis of Act-VAE.} We select hyperparameters near the elbow point to balance semantic compactness with reconstruction fidelity.}
    \label{fig:exp_actvae}
\end{figure}

\subsection{Cross-Modal Alignment via CLAP}
\label{sec:model_clap}
For the Act-VAE, we adopt the Transformer-based encoder-decoder architecture from~\cite{mld}, which is optimized for modeling long-horizon kinematic sequences. A critical aspect of this stage is to balance token expressiveness with the learning difficulty of the downstream VLM. Larger $N_q$ or $K$ improves reconstruction fidelity by increasing latent capacity, but it also increases the token sequence length or vocabulary size that the policy must model. \add{We therefore use a rate-distortion analysis over $N_q$ and $K$, with the rate $r$ defined as:}
\begin{equation}
    r = \frac{N_q \cdot \log_2(K)}{N_a \cdot D_a \cdot \log_2(\frac{\mathrm{R}}{\sqrt{\mathrm{MSE}}})},
\end{equation}
where $N_q$ is the latent sequence length, $K$ is the codebook size, $N_a$ is the action chunk size, $D_a$ is the action dimension, and $\mathrm{R}$ represents the dynamic range of the normalized action data. \add{The numerator measures the capacity of the discrete latent sequence, while the denominator estimates the information needed to represent the normalized continuous action chunk at the achieved reconstruction precision. Since actions are normalized to $[-1,1]$, $\sqrt{\mathrm{MSE}}$ approximates this precision. We also report $\mathrm{PSNR}=20\log_{10}(\mathrm{R}/\sqrt{\mathrm{MSE}})$ as the reconstruction-fidelity metric.} We analyze PSNR against varying compression levels (see Fig.~\ref{fig:exp_actvae}) and select hyperparameters near the elbow point to preserve high reconstruction fidelity while keeping the latent representation compact enough for token-space policy learning.

For VD-VAE training, we use two architectural choices to ensure robust dynamics learning. First, to mitigate the noise inherent in pixel-space supervision~\cite{bu2025univla,hafner2019learning}, we compute losses in feature space using patch-level embeddings extracted from DINOv3~\cite{dinov3}. Second, we employ a factorized attention mechanism: the inverse-dynamics encoder uses spatiotemporal attention to capture motion cues, while the forward-dynamics decoder uses spatial attention. This design significantly reduces the GPU memory footprint while preserving essential spatiotemporal relationships. We also use~\cite{chen2023disco} for a memory-efficient distributed contrastive loss implementation.

\begin{figure*}[t]
    \begin{center}
        \includegraphics[width=\textwidth]{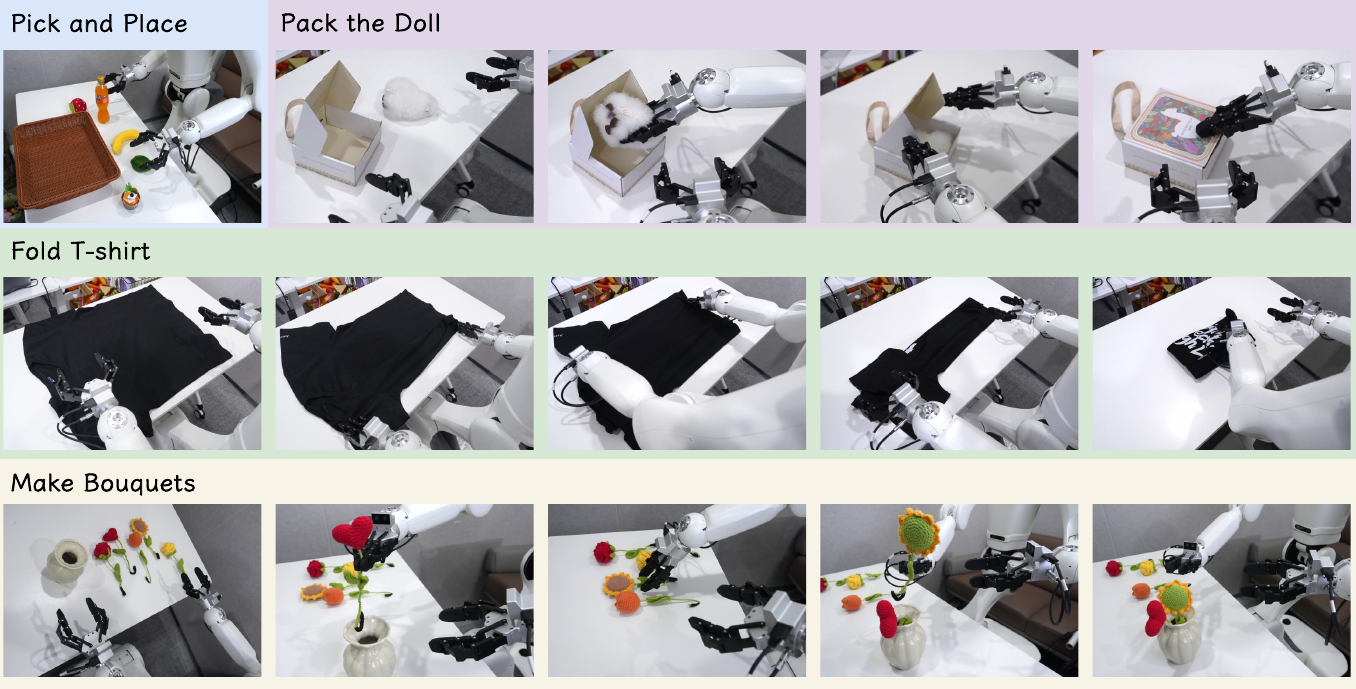}
    \end{center}
    \caption{\textbf{Visualization of the real-world deployment task processes.}}
    \label{fig:exp_real_vis}
\end{figure*}

\subsection{\texorpdfstring{\add{Policy Pretraining and Post-training}}{Policy Pretraining and Post-training}}
\label{sec:model_policy}
We implement our VLA models using Qwen3VL-4B~\cite{qwen3vl} as the foundational VLM, selected for its strong embodied reasoning capabilities. \add{The policy training schedule follows the method in Section~\ref{sec:method}: CLAP-NTP is trained in token space during pretraining, while RF and KM are used only during post-training and target-domain adaptation.}

\subsubsection{CLAP-NTP Training}
For the high-level planner, we adapt the Qwen3VL-4B tokenizer by initializing new tokens corresponding to the discrete action codebook $\mathcal{C}$ derived from Act-VAE. The model is trained using a \add{next-token prediction objective on robot action tokens and VD-VAE pseudo tokens from human videos} for a total of 150,000 steps. We use a peak learning rate of $5 \times 10^{-5}$ with a linear warmup over the first 1,000 steps. To ensure stable convergence, we apply a cosine decay schedule after 100,000 steps, reducing the learning rate to a minimum of $5 \times 10^{-6}$.

\subsubsection{\texorpdfstring{\add{CLAP-RF and KM Post-training}}{CLAP-RF and KM Post-training}}
\add{For target-domain post-training, we use the two-stage procedure in Algorithm~\ref{alg:km_ft}. In Stage 1, we fine-tune the NTP policy on all available tokenized target data, including robot tokens and human pseudo tokens, for three epochs over the dataset. This stage uses a learning rate of $2\times10^{-5}$ with a 1,000-step linear warmup and no learning-rate decay.}

The continuous action expert is trained using the Rectified Flow objective~\cite{rectified_flow}. To improve the model's robustness to noise, we sample the time step $t$ from the distribution $p(t)={\rm Beta}(\frac{s-t}{s};1.5, 1.0)$, following the methodology introduced in $\pi_0$~\cite{pi0}. \add{In Stage 2, we initialize the RF action head from the tuned NTP backbone and train the action expert with KM using only robot trajectories for five epochs over the dataset. The action expert uses a learning rate of $1\times10^{-4}$ with a 1,000-step linear warmup and cosine decay.}

To achieve faster inference, we make the action expert shallower than the VLM. Therefore, it cannot use all hidden features from the VLM. We found that the depth of feature extraction significantly affects performance. Empirically, fusing features from the early and middle layers of the VLM yields better results. This multi-scale feature aggregation allows the diffusion transformer to use both low-level visual details and mid-level semantic abstractions for precise action generation.

\begin{table*}[!t]
    \centering
    \renewcommand\arraystretch{1.6}
    \setlength\tabcolsep{3pt}
    \setlength{\aboverulesep}{0pt}
    \setlength{\belowrulesep}{0pt}
    \small
    \caption{\textbf{Detailed performance of CLAP and baselines in real-world tasks under the original setup.} \add{The \textbf{best} and \underline{second-best} results within each category are highlighted.}}
    \begin{tabular}{c|ccccccccc|c}
        \hline
        \multirow{2}{*}{\textbf{Method}}  & \multicolumn{2}{c}{\textit{\textbf{Pick and Place}}} & \multicolumn{2}{c}{\textit{\textbf{PnP (OOD)}}} & \multicolumn{2}{c}{\textit{\textbf{Pack the Doll}}} & \multicolumn{1}{c}{\textit{\textbf{Fold T-shirt}}} & \multicolumn{2}{c|}{\textit{\textbf{Make Bouquets}}} & \multirow{2}{*}{\textit{\textbf{Task Mean}}} \\
        \cmidrule(r){2-3}\cmidrule(r){4-5}\cmidrule(r){6-7}\cmidrule(r){8-8}\cmidrule(r){9-10}
        & \textit{Pick (\%)} & \textit{Place (\%)} & \textit{Pick (\%)} & \textit{Place (\%)} & \textit{P\&P (\%)} & \textit{Close (\%)} & \textit{Succ. (\%)} & \textit{C-1 (\%)} & \textit{C-2 (\%)} &  \\
        \hline
        \textit{$\pi_0$}~\cite{pi0} & 85 & 75 & 65 & 60 & \underline{80} & \underline{60} & \underline{40} & \textbf{40} & \underline{30} & 54.0 \\
        \textit{$\pi_{0.5}$}~\cite{pi05} & \underline{90} & \underline{80} & \underline{80} & \underline{75} & \underline{80} & \underline{60} & \textbf{50} & \underline{30} & \textbf{40} & \underline{60.0} \\
        \textit{UniVLA}~\cite{bu2025univla} & 75 & 60 & 65 & 50 & 70 & 30 & 10 & \underline{30} & 20 & 35.0 \\
        \rowcolor[HTML]{fde8eb}
        \textbf{\textit{CLAP-NTP}} & \underline{90} & \underline{80} & \textbf{85} & \textbf{85} & \underline{80} & \textbf{70} & 30 & \underline{30} & \underline{40} & 58.7 \\
        \rowcolor[HTML]{fde8eb}
        \textbf{\textit{CLAP-RF}} & \textbf{95} & \textbf{85} & \textbf{85} & 70 & \textbf{90} & \textbf{70} & \underline{40} & \textbf{40} & \textbf{50} & \textbf{62.7} \\ 
        \hline
    \end{tabular}
    \label{tab:real_result}
\end{table*}

\begin{figure*}[t]
    \begin{center}
        \includegraphics[width=\textwidth]{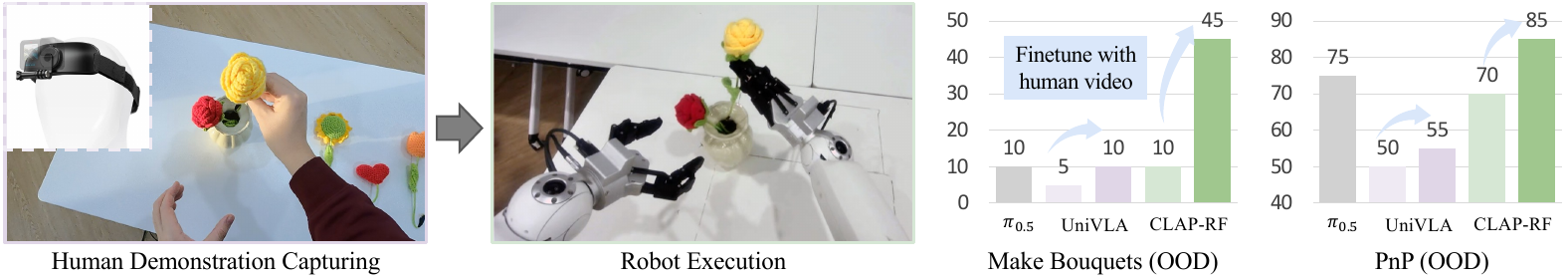}
    \end{center}
    \caption{\add{\textbf{Comparison of generalization capability when incorporating human egocentric video data.}}}
    \label{fig:exp_real_human}
\end{figure*}

\section{Evaluation}
\label{sec:exp}

In this section, we present an extensive evaluation of the proposed CLAP framework. We validate our method through experiments on both a real-world robotic platform and simulation environments using LIBERO~\cite{libero}. Beyond standard performance metrics, we analyze the learned latent action space to quantify the alignment between visual dynamics and physical control. Our evaluation addresses the following research questions:
\begin{enumerate}
    \item \textbf{Performance \& Precision:} Can CLAP-NTP and CLAP-RF effectively execute complex bimanual manipulation tasks, and does the hierarchical design enable high-precision control? We study this in Section~\ref{sec:exp_deploy}.
    \item \textbf{Generalizability:} Does the model robustly adapt to unseen objects, varying environmental conditions, and new embodiments? We evaluate this in Sections~\ref{sec:exp_deploy} and~\ref{sec:exp_sim}.
    \item \textbf{Cross-Modal Alignment:} How effective is the learned latent space in bridging the domain gap between human videos and robotic data? We analyze this in Section~\ref{sec:exp_human}.
\end{enumerate}

\subsection{Real-world Robot Deployment}
\label{sec:exp_deploy}

\subsubsection{Experimental Setup}
We conducted real-world experiments on the Astribot S1, a high-precision dual-arm robot. To match the pretraining data distribution, we lock the robot chassis and torso and control only the 14-DoF dual arms and grippers. The policy receives RGB streams from a head-mounted camera centered on the workspace and two wrist-mounted cameras.

\subsubsection{Task Design}
We designed five tasks to evaluate capabilities ranging from basic manipulation to semantic reasoning and deformable-object interaction. Fig.~\ref{fig:exp_real_vis} shows the task procedures.
\begin{enumerate}
    \item \textbf{Pick and Place (Seen):} Evaluates basic manipulation. We use 10 objects seen during pretraining and test each object in two trials, giving 20 episodes per model.
    \item \textbf{Pick and Place (OOD):} Tests generalization to novel geometries and textures. We use 10 objects unseen in the training data and run 20 trials per model.
    \item \textbf{Pack the Doll:} Requires long-horizon, multi-stage planning: picking up a doll, placing it precisely into a box, and closing the lid. This task tests precise geometric reasoning. We collected 200 teleoperated demonstrations for fine-tuning and evaluated each model over 10 trials.
    \item \textbf{Fold T-shirt:} Tests bimanual manipulation of deformable objects. Starting from a flat T-shirt, the robot must execute a folding sequence with coordinated dual-arm motion. We use 200 fine-tuning demonstrations and evaluate each policy over 10 trials.
    \item \textbf{Make Bouquets:} Tests instruction following and semantic grounding. The scene contains five wool flowers; the robot must identify two flowers specified by natural language and place them in a vase. We collected 100 demonstrations for each of two flower combinations and evaluated each model 10 times per combination.
\end{enumerate}

\subsubsection{Baselines}
We compare our approach with three strong baselines:
\begin{itemize}
    \item \textbf{$\pi_0$ and $\pi_{0.5}$}~\cite{pi0,pi05}: State-of-the-art generalist VLA policies trained on large public and private robotics datasets. They serve as strong references for large-scale transfer learning.
    \item \textbf{UniVLA}~\cite{bu2025univla}: A recent VLA approach that also uses latent action tokens. This comparison helps isolate the benefits of our CLAP alignment and hierarchical control strategy.
\end{itemize}

\begin{table*}[!t]
    \centering
    \renewcommand\arraystretch{1.6}
    \setlength\tabcolsep{3pt}
    \setlength{\aboverulesep}{0pt}
    \setlength{\belowrulesep}{0pt}
    \small
    \caption{\textbf{Results on robustness evaluations under environmental perturbations.} \add{The \textbf{best} and \underline{second-best} results within each category are highlighted.}}
    \begin{tabular}{l|cc|cccccc|c}
        \hline
        \multirow{2}{*}{\textbf{Method}} & \multicolumn{2}{c|}{\textbf{\textit{Original Setting}}} & \multicolumn{2}{c}{\textbf{\textit{Background Change}}} & \multicolumn{2}{c}{\textbf{\textit{Lighting Variation}}} & \multicolumn{2}{c|}{\textbf{\textit{Novel Object}}} & \multirow{2}{*}{\textit{\textbf{Mean}}} \\
        \cmidrule(r){2-3}\cmidrule(r){4-5}\cmidrule(r){6-7}\cmidrule(r){8-9}
        & \textit{P\&P (\%)} & \textit{Close (\%)} & \textit{P\&P (\%)} & \textit{Close (\%)} & \textit{P\&P (\%)} & \textit{Close (\%)} & \textit{P\&P (\%)} & \textit{Close (\%)} &  \\
        \hline
        \textit{$\pi_0$}~\cite{pi0} & \underline{80} & \underline{60} & \underline{70} & 50 & 60 & 40 & 60 & 50 & 46.7 \\
        \textit{$\pi_{0.5}$}~\cite{pi05} & \underline{80} & \underline{60} & \textbf{80} & \underline{60} & \textbf{80} & \underline{50} & \underline{70} & \underline{60} & \underline{56.7} \\
        \textit{UniVLA}~\cite{bu2025univla} & 70 & 30 & 60 & 20 & 50 & 10 & 50 & 20 & 16.7 \\
        \rowcolor[HTML]{fde8eb}
        \textbf{\textit{CLAP-RF}} & \textbf{90} & \textbf{70} & \textbf{80} & \textbf{70} & \underline{70} & \textbf{70} & \textbf{80} & \textbf{70} & \textbf{70.0} \\ 
        \hline
    \end{tabular}
    \label{tab:real_gen}
\end{table*}

\begin{figure}[t]
    \begin{center}
        \includegraphics[width=0.48\textwidth]{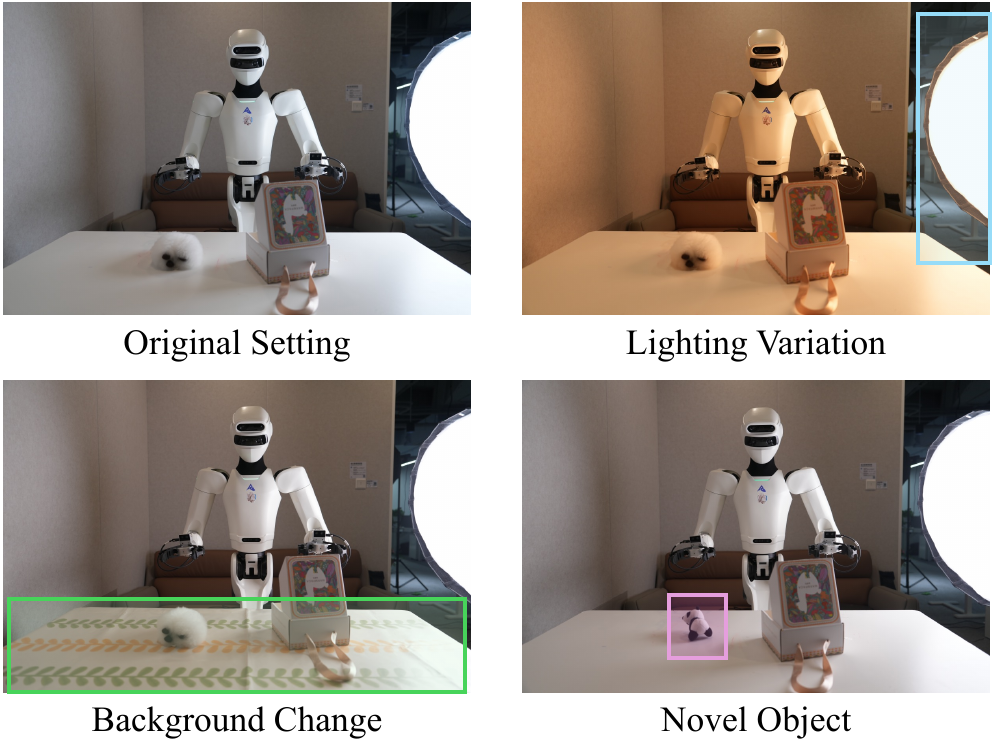}
    \end{center}
    \caption{\textbf{Setup for generalizability evaluations.}}
    \label{fig:exp_real_gen}
\end{figure}

\subsubsection{Results and Analysis}
Table~\ref{tab:real_result} summarizes the real-world evaluation. The results support four observations.

\bfs{CLAP-RF achieves strong performance on the platform.}
CLAP-RF achieves the highest mean success rate among the evaluated methods at 62.7\%, compared with $\pi_0$ at 54.0\% and $\pi_{0.5}$ at 60.0\%. This result supports the effectiveness of the post-trained RF controller under the evaluated real-world setting. Since CLAP differs from UniVLA in multiple aspects, including backbone, data recipe, alignment objective, and post-training strategy, we use the component ablations in Table~\ref{tab:ablation_clap_ntp_color} to isolate these factors rather than attributing the performance gap to a single component.

\bfs{Precision and planning across variants.}
Across the two variants, CLAP-RF performs better on tasks that require high precision. In \textit{Pack the Doll}, CLAP-RF improves the pick-and-place stage from 80\% to 90\% and matches CLAP-NTP on the tight-tolerance ``Close'' sub-task at 70\%. In \textit{Fold T-shirt}, which requires smooth bimanual coordination, CLAP-RF raises success from 30\% to 40\%. This supports our hypothesis that the discrete NTP model benefits high-level perception and reasoning, while the RF expert is better suited to complex dynamics and fine-grained motor control due to the continuous representation formulation.

\bfs{Robust generalization to OOD objects.}
On \textit{Pick and Place (OOD)}, CLAP-NTP maintains high performance, with 85\% success for both pick and place. This matches or exceeds its performance on seen objects, indicating that the visual encoder and aligned latent space learn generalized manipulability rather than memorizing object instances. CLAP-RF drops slightly on OOD placement to 70\%, suggesting that the continuous diffusion policy is more sensitive to visual distribution shifts than the discrete token predictor, while still remaining competitive.

\bfs{Semantic understanding and instruction following.}
\textit{Make Bouquets} stresses language-conditioned object selection. Both CLAP-NTP and CLAP-RF perform strongly, reaching up to 50\% success and matching or exceeding the large-scale $\pi_0$ and $\pi_{0.5}$ baselines.

\begin{table}[t]
    \centering
    \caption{\add{\textbf{Failure-case analysis on OOD real-world tasks.} We categorize failed trials into three mutually exclusive failure modes over 20 attempts per setting: \textit{attempt failure}, where the policy attempts to grasp an incorrect target object; \textit{pick failure}, where the correct object is targeted but the grasp or lift is unsuccessful; and \textit{place failure}, where the object is picked but not placed at the specified goal location.}}
    \label{tab:failure_cases}
    \resizebox{\linewidth}{!}{
        \begin{tabular}{l|ccc|ccc}
            \toprule
            \multirow{2}{*}{\textbf{Setting}} & \multicolumn{3}{c|}{\textbf{PnP (OOD)}} & \multicolumn{3}{c}{\textbf{Make Bouquets (OOD)}} \\
            \cmidrule(lr){2-4} \cmidrule(lr){5-7}
            & \textbf{Attempt} & \textbf{Pick} & \textbf{Place} & \textbf{Attempt} & \textbf{Pick} & \textbf{Place} \\
            \midrule
            \textit{w/ Human Data} & \add{1} & \add{2} & \add{0} & \add{0} & \add{2} & \add{9} \\
            \textit{w/o Human Data} & \add{4} & \add{2} & \add{0} & \add{8} & \add{3} & \add{7} \\
            \bottomrule
        \end{tabular}
    }
\end{table}

Overall, the real-world experiments show that CLAP adapts VLMs to physical robot control, with CLAP-NTP excelling in instruction following and CLAP-RF providing the precision needed for complex, contact-rich manipulation.

\subsubsection{Generalization via Human Demonstrations}
\label{sec:exp_human}

We further evaluate whether the shared latent action space in Section~\ref{sec:method_clap} enables the model to use human video demonstrations for object generalization.

\bfs{Experimental Design.}
We use \textit{Make Bouquets (OOD)} \add{and \textit{PnP (OOD)} as object-generalization testbeds}. For \textit{Make Bouquets}, the initial teleoperation dataset contains only two flower combinations, such as ``red heart and yellow sunflower.'' Preliminary experiments showed that policies trained only on these data overfit severely and failed to generalize to novel combinations such as ``orange tulip and red rose.''

Human-video collection follows the same egocentric protocol across the OOD tasks. We used a head-mounted GoPro 9 to capture video from a viewpoint close to the robot head camera. During collection, the human operator used their hands to mimic the robot gripper, performing simple open and close motions while avoiding complex grasping dynamics. Fig.~\ref{fig:exp_real_human} shows examples. For \textit{Make Bouquets}, we collected three additional settings with 100 episodes each, covering all five seen flower types. \add{For \textit{PnP (OOD)}, we used the same protocol and collected 20 human-video trajectories for each OOD object.}\footnote{Video data collected for this study were fully anonymized and contained no personally identifiable information.}

\bfs{Comparative Analysis.}
We compare CLAP-RF with $\pi_{0.5}$ and UniVLA.
\begin{itemize}
    \item \textbf{$\pi_{0.5}$:} Trained only on teleoperation data.
    \item \textbf{UniVLA:} For a fair comparison, we trained UniVLA with its provided visual tokenizer and then fine-tuned it with an additional action head using teleoperation data.
    \item \textbf{CLAP-RF:} Evaluated before and after fine-tuning with pseudo-labeled human videos generated by our VD-VAE.
\end{itemize}

\bfs{Results.}
Fig.~\ref{fig:exp_real_human} presents the results. On \textit{Make Bouquets (OOD)}, models trained only on teleoperation data overfit to the training distribution, and all models stayed at or below 10\% success on unseen flower collections.

After fine-tuning with human data, CLAP-RF improves from 10\% to 45\% on \textit{Make Bouquets (OOD)} \add{and from 70\% to 85\% on \textit{PnP (OOD)}}. In contrast, UniVLA improves only from 5\% to 10\% on \textit{Make Bouquets (OOD)} \add{and from 50\% to 55\% on \textit{PnP (OOD)}}. These results support our claim that CLAP's alignment mechanism enables effective transfer of semantic object knowledge from unlabeled human videos to robotic control.

\bfs{Failure-Case Analysis.}
\add{To separate semantic grounding errors from execution failures, we categorize unsuccessful OOD trials in Table~\ref{tab:failure_cases} for settings with and without human video data. All counts are computed over 20 attempts per setting. An \textit{attempt failure} means that the policy initiates a grasp toward the wrong object, directly reflecting object-selection or language-grounding errors. A \textit{pick failure} means that the policy selects the intended target but fails to stably grasp or lift it. A \textit{place failure} occurs after a successful pick when the object is not delivered to the specified goal region.}

\add{The failure distribution confirms that human video data provides effective semantic supervision. With human data, attempt failures decrease from 4 to 1 on OOD Pick-and-Place and from 8 to 0 on OOD Make Bouquets. These reductions indicate that human videos help the policy choose the correct object and follow the instruction across unseen object and task combinations.}

\subsubsection{Robustness Evaluation}
To evaluate robustness under environmental perturbations, we conducted stress tests with three variations shown in Fig.~\ref{fig:exp_real_gen}: \textbf{Background Change}, which introduces a patterned tablecloth; \textbf{Lighting Variation}, which changes illumination intensity and color temperature; and \textbf{Novel Object}, which replaces the target with an unseen instance or introduces distractors.

As shown in Table~\ref{tab:real_gen}, \textbf{CLAP-RF} achieves the best robustness among the evaluated methods, with a mean success rate of \textbf{70.0\%}, compared with $\pi_{0.5}$ at 56.7\% and UniVLA at 16.7\%. CLAP-RF remains strong under background shifts, reaching 80\% on Pick \& Place and 70\% on Close. Together with the ablation in Table~\ref{tab:ablation_clap_ntp_color}, this robustness is consistent with CLAP reducing sensitivity to action-irrelevant visual factors. UniVLA is more brittle to these shifts, likely because its reconstruction-based objective encodes extraneous visual details. Although $\pi_{0.5}$ remains competitive under lighting variation due to its large-scale pretraining, CLAP-RF performs better on the precision-heavy Close task, with 70\% success versus 50\%.

\begin{table}[t]
    \centering
    \caption{\textbf{Results on the LIBERO benchmark.} We compare success rates (\%) across different evaluation suites. The table is categorized into methods that train separate models for each suite (top) and generalist models trained once across all suites (bottom). The \textbf{best} and \underline{second-best} results in each column are highlighted. Note that $^*$LAPA results are reproduced by UniVLA authors using Prismatic-7B, and $\pi_0$ (PaliGemma) is initialized from PaliGemma-3B~\cite{paligemma} without VLA pretraining.}
    \label{tab:libero_combined_v3}
    \resizebox{0.48\textwidth}{!}{
        \begin{tabular}{l|cccc|c}
            \toprule
            \textbf{Method} & \textbf{Spatial} & \textbf{Object} & \textbf{Goal} & \textbf{Long} & \textbf{\textit{Average}} \\
            \midrule

            \multicolumn{6}{l}{\textit{\footnotesize Separate models for each task suite}} \\
            % -----------------------
            
            LAPA$^*$ \cite{lapa} & 73.8 & 74.6 & 58.8 & 55.4 & 65.7 \\
            Diffusion Policy \cite{diffusionpolicy} & 78.3 & 92.5 & 68.3 & 50.5 & 72.4 \\
            Octo \cite{octo_2023} & 78.9 & 85.7 & 84.6 & 51.1 & 75.1 \\
            OpenVLA (7B) \cite{openvla} & 84.7 & 88.4 & 79.2 & 53.7 & 76.5 \\
            UniVLA~\cite{bu2025univla} & 96.5 & 96.8 & 95.6 & 92.0 & 95.2 \\
            \add{OpenVLA-OFT~\cite{oftvla}} & \add{97.6} & \add{98.4} & \add{\underline{97.9}} & \add{94.5} & \add{97.1} \\
            \midrule
            
            \multicolumn{6}{l}{\textit{\footnotesize Generalist models trained once}} \\
            % -----------------------
            
            SmolVLA~\cite{smolvla} & 93.0 & 94.0 & 91.0 & 77.0 & 88.8 \\
            \add{FLOWER~\cite{reuss2025flower}} & \add{97.5} & \add{\underline{99.1}} & \add{96.1} & \add{\underline{94.9}} & \add{96.9} \\
            $\pi_0$ (PaliGemma)~\cite{pi0} & 87.0 & 63.0 & 89.0 & 48.0 & 71.8 \\
            $\pi_0$~\cite{pi0} & 90.0 & 86.0 & 95.0 & 73.0 & 86.0 \\
            $\pi_{0.5}$~\cite{pi05} & \textbf{98.8} & 98.2 & \textbf{98.0} & 92.4 & 96.9 \\
            \add{X-VLA~\cite{xvla}} & \add{98.2} & \add{98.6} & \add{97.8} & \add{\textbf{97.6}} & \add{\textbf{98.1}} \\
            \midrule
            \rowcolor[HTML]{fde8eb}
            \add{\textbf{CLAP-RF}} & \add{\underline{98.6}} & \add{\textbf{99.2}} & \add{\textbf{98.0}} & \add{93.0} & \add{\underline{97.2}} \\
            \bottomrule
        \end{tabular}
    }
\end{table}

\subsection{Simulation Results}
\label{sec:exp_sim}

\bfs{Experiment Setup.} To evaluate our method in a controlled environment, we use the LIBERO benchmark~\cite{libero}, a standard suite designed for lifelong robotic learning. Our evaluation focuses on supervised fine-tuning, where policies are trained via behavioral cloning on expert demonstrations. The benchmark consists of four distinct task suites, each containing 10 tasks with 50 human-teleoperated demonstrations per task:
\begin{itemize}
    \item \textbf{LIBERO-Spatial:} Tests the agent's ability to reason about spatial relationships and geometric configurations, such as precise placement.
    \item \textbf{LIBERO-Object:} Evaluates generalization across different object instances while maintaining consistent scene layouts.
    \item \textbf{LIBERO-Goal:} Challenges the agent with diverse task objectives within consistent layouts, assessing goal-conditioned adaptability.
    \item \textbf{LIBERO-Long:} Focuses on long-horizon, multi-stage manipulation tasks, requiring complex planning across heterogeneous objects and layouts.
\end{itemize}

Following the protocol established in OpenVLA~\cite{openvla}, we filter out failure cases from the training data. We adopt a challenging \textit{generalist} training setting: rather than training separate experts for each suite, we train a single CLAP-RF policy across all four task subsets simultaneously. The model is fine-tuned for \add{30k} steps with a batch size of 128. We employ our proposed Knowledge Matching algorithm during fine-tuning to preserve pretrained priors. We report the average success rate over \add{500 trials per task suite, with 50 trials per task}.

\bfs{Baselines.} We compare our approach against state-of-the-art methods, categorized into two groups based on their training paradigms as shown in Table III:
\begin{itemize}
    \item \textit{Specialist Models:} These methods train separate models for each task suite, simplifying the learning problem. These baselines include LAPA~\cite{lapa}, Diffusion Policy~\cite{diffusionpolicy}, Octo~\cite{octo_2023}, OpenVLA~\cite{openvla}, UniVLA~\cite{bu2025univla}, and \add{OpenVLA-OFT~\cite{oftvla}}.
    \item \textit{Generalist Models:} These methods, like ours, train a single model across all suites, requiring the policy to handle diverse distributions simultaneously. These baselines include the PaliGemma-based $\pi_0$~\cite{pi0}, the full $\pi_0$~\cite{pi0}, $\pi_{0.5}$~\cite{pi05}, SmolVLA~\cite{smolvla}, \add{X-VLA~\cite{xvla}}, and \add{FLOWER~\cite{reuss2025flower}}.
\end{itemize}

\bfs{Results.} The quantitative results on the LIBERO benchmark are summarized in Table~\ref{tab:libero_combined_v3}. \add{CLAP-RF achieves a strong average success rate of \textbf{97.2\%} in the generalist setting, outperforming SmolVLA at 88.8\%, $\pi_0$ at 86.0\%, $\pi_{0.5}$ at 96.9\%, and FLOWER at 96.9\%, while remaining close to X-VLA at 98.1\%. It also slightly exceeds the specialist OpenVLA-OFT baseline at 97.1\% despite using a single policy across all four suites.}

\begin{table}[t]
    \centering
    \small
    \caption{\textbf{Rate-distortion analysis of Act-VAE.} We evaluate the trade-off between semantic compactness and reconstruction fidelity by varying the latent sequence length ($N_q$) and codebook size ($K$). The \textbf{selected configuration} (highlighted) balances high reconstruction quality (PSNR) with a compact token rate ($r$).}
    \label{tab:act_vae}
    \label{tab:act_vae_ablation}
    \resizebox{0.4\textwidth}{!}{
        \begin{tabular}{c|c|ccc}
            \toprule
            \textbf{$N_q$} & \textbf{$K$} & \textbf{MSE} & \textbf{PSNR (dB)} & \textbf{Rate ($r$)} \\
            \midrule
            
            12 & 128 & 0.0023 & 32.40 & 0.070 \\
            12 & 256 & 0.0010 & 36.02 & 0.072 \\
            12 & 512 & 0.0007 & 37.57 & 0.077 \\
            
            \midrule
            
            36 & 256 & 0.0002 & 43.01 & 0.175 \\
            20 & 256 & 0.0003 & 41.25 & 0.104 \\
            
            % Selected Parameter Row (Red Dot)
            \rowcolor[HTML]{fde8eb}
            \textbf{16} & \textbf{256} & \textbf{0.0004} & \textbf{40.00} & \textbf{0.086} \\
            
            8 & 256 & 0.0041 & 29.89 & 0.058 \\
            4 & 256 & 0.1022 & 15.93 & 0.054 \\
            
            \bottomrule
        \end{tabular}
    }
\end{table}

\begin{table}[t]
    \centering
    \caption{\textbf{\add{Post-training strategy ablation across simulation and real-world tasks.}} \add{We evaluate feature selection and fine-tuning strategies on the four LIBERO suites and Pick-and-Place (PnP) tasks under ID and OOD object settings.}}
    \label{tab:ablation_libero}
    \resizebox{0.48\textwidth}{!}{
        \begin{tabular}{l|ccccc|cc}
            \toprule
            \multirow{2}{*}{\textbf{Configuration}} & \multicolumn{5}{c|}{\textbf{LIBERO}} & \multicolumn{2}{c}{\add{\textbf{PnP}}} \\
            \cmidrule(lr){2-6}\cmidrule(lr){7-8}
            & \textbf{Spatial} & \textbf{Object} & \textbf{Goal} & \textbf{Long} & \textbf{\textit{Avg.}} & \add{\textbf{ID}} & \add{\textbf{OOD}} \\
            \midrule
            \multicolumn{8}{l}{\textit{\textbf{Feature Level Analysis}}} \\
            \hspace{3mm} w/ high-level feats & 94.8 & 97.5 & 91.3 & 89.9 & 93.4 & \add{70} & \add{65} \\
            \hspace{3mm} w/ low-level feats  & 97.2 & 98.8 & 97.6 & 90.6 & 96.1 & \add{75} & \add{60} \\

            \midrule
            \multicolumn{8}{l}{\textit{\textbf{Fine-tuning Strategy Analysis}}} \\
            \hspace{3mm} w/ KI~\cite{ki} & 90.4 & 97.4 & 85.6 & 79.8 & 88.3 & \add{75} & \add{65} \\
            \hspace{3mm} ft. VLM & 97.9 & 99.2 & 97.6 & 90.0 & 96.2 & \add{85} & \add{60} \\

            \midrule
            \multicolumn{8}{l}{\textit{\textbf{Full Configuration}}} \\
            \hspace{3mm} low+mid-level feats + KM & 98.6 & 99.2 & 98.0 & 93.0 & 97.2 & \add{85} & \add{70} \\
            \bottomrule
        \end{tabular}
    }
\end{table}

\subsection{Ablation Study}
\label{sec:ablation}

We conduct ablation studies to validate the architectural decisions of our framework, focusing on the quantization dynamics of Act-VAE and the structural strategies of the CLAP-RF policy.

\subsubsection{Rate-Distortion Trade-off in Act-VAE}
We analyze the trade-off between semantic compactness and reconstruction fidelity through an information-theoretic lens. The information capacity of a latent trajectory is governed by the product $N_q \cdot \log(K)$. Reconstruction quality measured by PSNR is positively correlated with this capacity, as high-frequency motion details, which typically contain greater information density, require a larger latent space to be accurately preserved.

As detailed in Table~\ref{tab:act_vae}, increasing $N_q$ or $K$ naturally improves PSNR but also increases the latent token capacity that the downstream VLM must model. For the policy, learning difficulty scales with both the sequence length $N_q$ and vocabulary size $K$. Excessive sequence lengths or vocabulary sizes dilute attention and hinder the model's ability to capture semantic dependencies. Consequently, we aim to maximize fidelity without sacrificing the compactness required for effective VLM training. We identify the highlighted configuration $N_q=16, K=256$ as the optimal ``elbow point''. \add{This setting preserves high reconstruction fidelity while keeping the latent representation compact enough for token-space policy learning.}

\subsubsection{\add{Backbone and Data Recipe}}
\begin{table}[t]
    \centering
    \small
    \renewcommand\arraystretch{1.12}
    \setlength\tabcolsep{4pt}
    \caption{\textbf{Ablation study on cross-modal alignment, backbone, and data sources.} We evaluate the impact of the backbone, data recipe, pseudo-positive labels, contrastive alignment loss, and inclusion of human video data on in-distribution (ID) and out-of-distribution (OOD) generalization. Performance is reported as success rates (\%) on real-world tasks.}
    \label{tab:ablation_clap_ntp_color}
    \resizebox{\linewidth}{!}{
        \begin{tabular}{l|cc|cc|c}
            \toprule
            \multirow{2}{*}{\textbf{Method}} &
            \multicolumn{2}{c|}{\textbf{Pick \& Place}} &
            \multicolumn{2}{c|}{\textbf{Make Bouquets}} &
            \multirow{2}{*}{\textbf{\textit{Average}}} \\
            \cmidrule(lr){2-3} \cmidrule(lr){4-5}
            & \textit{ID} & \textit{OOD} & \textit{ID} & \textit{OOD} & \\
            \midrule

            \multicolumn{6}{l}{\add{\textit{\textbf{Backbone and Data Recipe}}}} \\
            \add{\textit{UniVLA baseline}~\cite{bu2025univla}} & \add{60} & \add{50} & \add{25} & \add{10} & \add{36.3} \\
            \add{\hspace{3mm} + Qwen3VL} & \add{65} (\textcolor{green!50!black}{+5}) & \add{60} (\textcolor{green!50!black}{+10}) & \add{25} (\textcolor{gray}{+0}) & \add{10} (\textcolor{gray}{+0}) & \add{40.0} (\textcolor{green!50!black}{+3.7}) \\
            \add{\hspace{3mm} + our data recipe} & \add{75} (\textcolor{green!50!black}{+15}) & \add{65} (\textcolor{green!50!black}{+15}) & \add{30} (\textcolor{green!50!black}{+5}) & \add{10} (\textcolor{gray}{+0}) & \add{45.8} (\textcolor{green!50!black}{+9.5}) \\
            \midrule

            \multicolumn{6}{l}{\add{\textit{\textbf{Contrastive Learning and Human Data}}}} \\
            \textbf{CLAP-NTP (Full)} & 90 & 80 & 35 & 35 & 60.0 \\
            \add{\hspace{3mm} w/o Pseudo Positive Label} & \add{85} (\textcolor{red}{-5}) & \add{80} (\textcolor{gray}{0}) & \add{35} (\textcolor{gray}{0}) & \add{35} (\textcolor{gray}{0}) & \add{58.8} (\textcolor{red}{-1.2}) \\
            \hspace{3mm} w/o Contrastive & 85 (\textcolor{red}{-5}) & 75 (\textcolor{red}{-5}) & 35 (\textcolor{gray}{0}) & 20 (\textcolor{red}{-15}) & 53.8 (\textcolor{red}{-6.2}) \\
            \hspace{3mm} w/o Human Data & 80 (\textcolor{red}{-10}) & 75 (\textcolor{red}{-5}) & 30 (\textcolor{red}{-5}) & 5 (\textcolor{red}{-30}) & 47.5 (\textcolor{red}{-12.5}) \\
            \add{\hspace{3mm} w/ Prismatic-7B~\cite{prismaticvlms}} & \add{90} (\textcolor{gray}{0}) & \add{75} (\textcolor{red}{-5}) & \add{35} (\textcolor{gray}{0}) & \add{25} (\textcolor{red}{-10}) & \add{56.3} (\textcolor{red}{-3.7}) \\
            \bottomrule
        \end{tabular}
    }
\end{table}

We perform an ablation study on CLAP-NTP in Table~\ref{tab:ablation_clap_ntp_color} to validate the effects of the model backbone, data recipe, pseudo-positive labels, contrastive alignment, and human video data.

\bfs{\add{Backbone and training recipe.}} \add{We first isolate the effect of the model backbone and data recipe. Replacing the UniVLA-style Prismatic-7B backbone with Qwen3VL improves the average success rate from 36.3\% to 40.0\%, even though Qwen3VL is smaller, suggesting that the newer backbone contributes stronger visual-language representations. Incorporating our data recipe further improves the average success rate to 45.8\%, with clear gains on Pick-and-Place ID/OOD settings. This indicates that the richer and more diverse training mixture improves object-level and OOD manipulation robustness. Nevertheless, these factors alone do not explain the full performance: CLAP-NTP (Full) reaches 60.0\%, and CLAP-NTP with the UniVLA-style Prismatic-7B backbone still achieves 56.3\%. These results show that modern backbones and diverse data are beneficial, but the proposed CLAP training mechanism remains the primary source of the remaining improvement.}

\subsubsection{Contrastive Learning and Human Data}
To validate the effects of the pseudo-positive labels, contrastive alignment, and human video data, we also perform an ablation study on the key components of our method.

\bfs{\add{Contrastive alignment.}} \add{Removing the pseudo-positive labels causes a modest drop from 60.0\% to 58.8\%, indicating that the EMA-based target provides useful alignment supervision for human-video transitions. Removing the full contrastive objective causes a larger degradation to 53.8\%, especially on Make Bouquets (OOD), confirming that contrastive alignment is important for shaping the latent distribution and improving generalization.}

\bfs{\add{Human video data.}} \add{Excluding human video data causes severe degradation, dropping the average success rate by 12.5\%. Make bouquets (OOD) performance collapses to 5\%, confirming that large-scale human data is indispensable for semantic generalization beyond robotic data.}

\subsubsection{Post-Training Strategy}
We further evaluate post-training strategies across both the LIBERO benchmark and real-world Pick-and-Place tasks in Table~\ref{tab:ablation_libero}.

\bfs{Feature selection.} Using high-level features alone gives 93.4\% average success on LIBERO, while low-level features improve the average to 96.1\%. This suggests that continuous action generation benefits more from geometry- and motion-sensitive representations than from purely semantic embeddings. The full model further combines low-level and mid-level features, so the RF head can use both local control cues and intermediate semantic abstractions.

\bfs{Post-training strategy.} KI is too restrictive for target-domain adaptation, yielding only 88.3\% average success on LIBERO. Direct VLM fine-tuning improves LIBERO performance to 96.2\%, but reduces PnP OOD success to 60\%, indicating weaker generalization after unconstrained updates. In contrast, the Full configuration reaches 97.2\% on LIBERO and 70\% on PnP OOD. These results show that KM improves target-domain adaptation while preserving the generalization ability of the pretrained token policy.

\subsection{More Analysis}
\label{sec:exp_analysis}
\bfs{Action latent space.}
To qualitatively validate the alignment between visual dynamics and physical control, we visualize retrieved video clips corresponding to learned latent representations in Fig.~\ref{fig:latent_act}. Given the high dimensionality and diversity of the 256-entry codebook, exact token matches across heterogeneous datasets are sparse. We therefore cluster the action tokens into 32 semantic groups and visualize samples from the same cluster. As shown, the learned latent space exhibits strong semantic consistency across domains. For instance, Group 1 captures the ``move right'' primitive, while Group 2 captures ``put down'', regardless of whether the agent is a human from Ego4D or a robot from Astribot/AgiBot. To verify that these latents encode precise motion rather than only high-level semantics, we decode the latent codes back into 3D trajectories using the action decoder. We project these 3D points onto the 2D image plane, visualized as red arrows in the Astribot S1 frames. The tight alignment between the projected arrows and the actual object manipulation confirms that our contrastive pretraining effectively grounds visual changes in physically executable actions. We visualize trajectories only for the self-collected Astribot dataset, as the accurate camera extrinsics required for 3D-to-2D projection were unavailable for the AgiBot and Ego4D datasets.

\begin{figure}[t]
    \begin{center}
        \includegraphics[width=\linewidth]{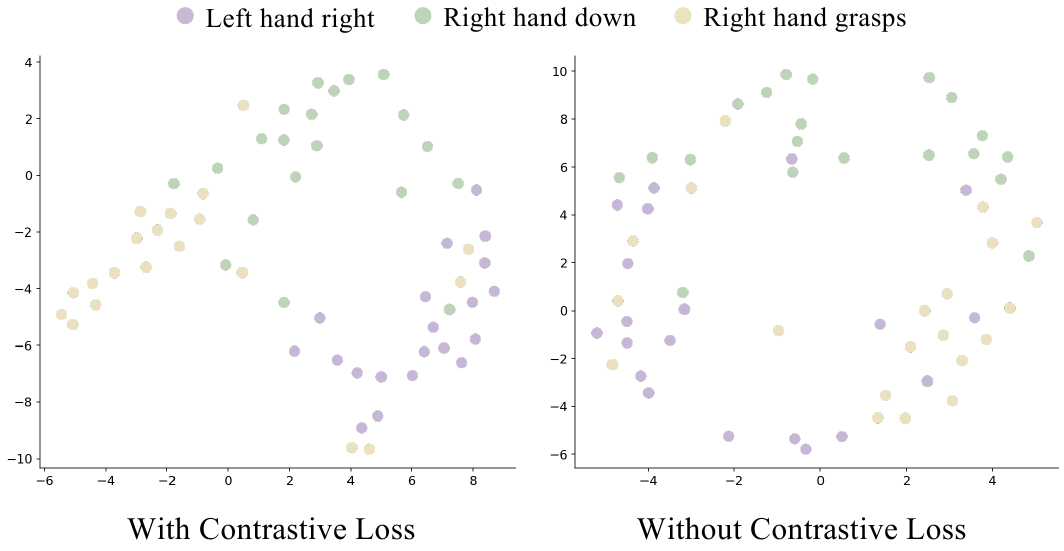}
    \end{center}
    \caption{\add{\textbf{t-SNE visualization of VD-VAE action features.} We manually select three semantic action categories, with 20 matched video pairs per category, and encode them using VD-VAE. Compared with the variant trained without the contrastive loss, CLAP produces more compact feature clusters for actions with the same semantics.}}
    \label{fig:tsne_vis}
\end{figure}

\add{We further compare the learned VD-VAE feature distribution with and without contrastive learning in Fig.~\ref{fig:tsne_vis}. Following the teaser visualization, we manually select three semantic action categories, each containing 20 paired examples, and apply t-SNE to the encoded features. With contrastive learning, features from the same semantic action form more compact clusters, whereas removing the contrastive loss leads to more dispersed and overlapping distributions. This indicates that the contrastive objective improves the organization of the action latent space beyond reconstructing visual dynamics alone.}

\begin{figure}[t]
    \begin{center}
        \includegraphics[width=0.48\textwidth]{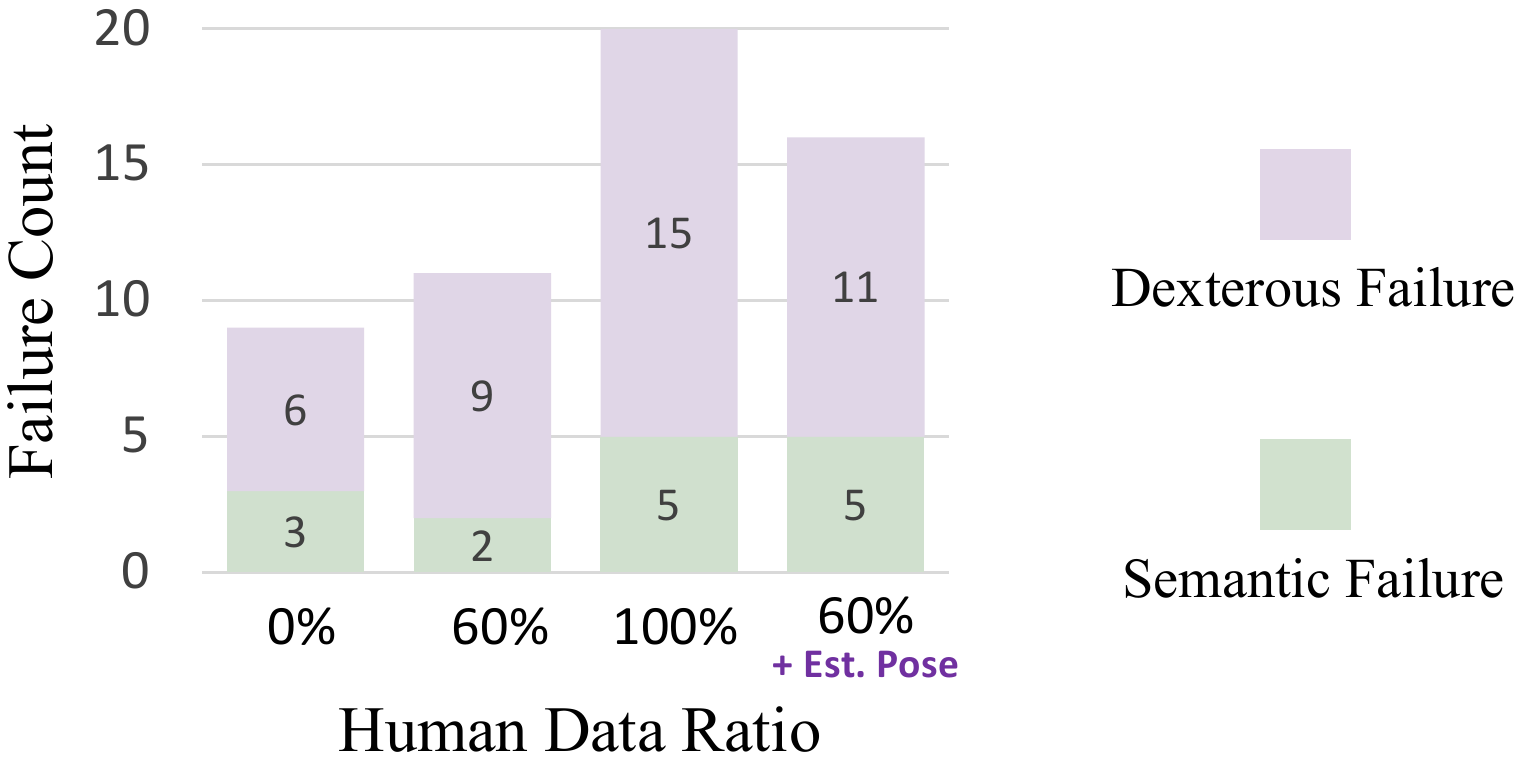}
    \end{center}
    \caption{\add{\textbf{Human-data and estimated-pose ablation on Make Bouquets.} We fix the training set to 500 episodes and vary the proportion of human videos, including an estimated-pose variant. Semantic failures reflect task or object-selection errors, whereas dexterous failures reflect inaccurate grasping or placement motions.}}
    \label{fig:failure_abl}
\end{figure}

\bfs{\add{Human data.}}
\add{To better understand whether human videos mainly transfer task semantics or low-level dexterous actions, we conduct a controlled ablation on the \textit{Make Bouquets} task. The total training budget is fixed to 500 episodes, and we vary the proportion of human videos in this mixture, with the remaining episodes drawn from robot teleoperation. We further construct an estimated-pose baseline using WiLoR~\cite{wilor}, an off-the-shelf monocular 3D hand localization and reconstruction method. For this baseline, the palm position is used as a proxy for the end-effector position, and the distance between the index fingertip and thumb fingertip is linearly mapped to the gripper width.}

\add{The results in Fig.~\ref{fig:failure_abl} show that human videos effectively reduce semantic failures, indicating that they help the policy learn object selection and instruction-conditioned task structure. However, without additional localization or tracking hardware, directly converting human videos into robot actions through estimated hand pose remains unreliable for embodiment-specific dexterous control. The estimated-pose variant introduces substantial localization noise and can even degrade the policy, as reflected by increased failures. This supports our design choice: since CLAP is intended to exploit large-scale unlabeled internet videos where accurate hardware-instrumented hand poses are unavailable, learning latent actions from visual dynamics provides a more scalable supervision signal than directly regressing robot end-effector trajectories from noisy human pose estimates.}

\bfs{Inference speed.}
Real-time responsiveness is essential for dynamic manipulation. We benchmark the inference latency of our models against representative baselines on a single \add{NVIDIA H100 GPU} using the LIBERO dataset; see Table~\ref{tab:infer_speed}. \add{For a fair comparison, all models are evaluated with a single $224 \times 224$ image as input.} The autoregressive CLAP-NTP model, while strong in reasoning, exhibits a higher latency of \add{382.0 ms} due to the sequential nature of token generation. In contrast, CLAP-RF reduces latency to \add{70.1 ms}. This performance is \add{slower than the highly optimized and smaller $\pi_0$ at 45.8 ms, but substantially faster than OpenVLA at 219.8 ms and FAST at 412.8 ms.}

\begin{table}[htb]
    \centering
    \caption{\textbf{Inference speed and GPU memory comparison.} All results are measured on a single \add{NVIDIA H100} with one $224 \times 224$ input image.}
    \small
    \resizebox{0.45\textwidth}{!}{
        \begin{tabular}{l|ccc}
        \toprule
            Method & \# params. (B) & Latency (ms) & Memory (G)\\ 
        \midrule
            $\pi_0$~\cite{pi0}      & 3.5 & 45.8 & 7.6 \\
            FAST~\cite{fast}        & 3.0 & 412.8 & 7.5 \\
            OpenVLA~\cite{openvla}  & 7.5 & 219.8 & 14.8 \\
        \midrule
            CLAP-NTP                & 4.5 & 382.0 & 9.2 \\
            CLAP-RF                 & 6.0 & 70.1 & 12.1 \\
        \bottomrule
        \end{tabular}
    }
\label{tab:infer_speed}
\end{table}

\section{Conclusion}

In this work, we addressed the challenge of data scarcity in robotic manipulation by using large-scale, unlabeled human video demonstrations. We identified that existing Latent Action Models often suffer from visual entanglement, where learned representations capture extraneous visual noise rather than manipulation skills. To overcome this, we proposed Contrastive Latent Action Pretraining (CLAP), a framework that explicitly aligns the visual latent space derived from human videos with a physically executable latent action space derived from robot trajectories. By enforcing this isomorphism through contrastive learning, we ensure that visual transitions are mapped to a quantized codebook grounded in physical control.

Building on these aligned representations, we introduced a dual-formulation VLA framework comprising CLAP-NTP, an autoregressive planner for semantic reasoning and instruction following, and CLAP-RF, a Rectified Flow-based policy designed for low-latency inference. In addition, our proposed Knowledge Matching (KM) regularization strategy mitigates catastrophic forgetting during fine-tuning. Extensive experiments across real-world bimanual tasks and the LIBERO simulation benchmark demonstrate that CLAP achieves competitive performance among baselines, enabling robust object generalization and precise control through the transfer of human visual priors.

Despite these advances, several limitations remain and suggest directions for future research. First, while CLAP successfully generalizes to novel objects within known tasks, generalizing to entirely new tasks solely from human videos remains a significant challenge. The current alignment captures high-level planning logic but may struggle to infer precise local dynamics for unseen activities without at least some robotic grounding. Second, the morphological discrepancy between human hands and robotic grippers introduces inherent ambiguity in the latent space. Although our contrastive approach aligns these modalities, complex dexterous human motions do not always have a direct mapping to parallel-jaw gripper actions, potentially limiting performance in fine-grained manipulation. \add{In our flower-manipulation human videos, the demonstrator intentionally used hand motions that resemble a parallel-jaw gripper, which reduces this ambiguity but also indicates that broader dexterous hand motions remain an important limitation.} Finally, our framework relies on a multi-stage training pipeline involving separate training for the VQ-VAEs, contrastive alignment, and policy heads. Future work will focus on unifying these stages into an end-to-end learning paradigm to reduce engineering complexity and further improve the efficiency of cross-embodiment transfer.

% Required packages in preamble:
% \usepackage{booktabs}
% \usepackage{amsmath}

\begin{table}[h]
    \centering
    \caption{\textbf{Model and training hyperparameters.} Training time is estimated using a single NVIDIA A100 80G GPU.}
    \label{tab:params}
    \resizebox{0.85\linewidth}{!}{
    \begin{tabular}{lc}
    \toprule
    \textbf{Hyperparameter} & \textbf{Value} \\
    \midrule
    \multicolumn{2}{l}{\textit{\textbf{Global Settings}}} \\
    \hspace{1em}Action Chunk Size & 32 \\
    
    \midrule
    \multicolumn{2}{l}{\textit{\textbf{Act-VAE}}} \\
    \hspace{1em}Total Training Steps & 100,000 \\
    \hspace{1em}VAE Learning Rate & $2 \times 10^{-5}$ \\
    \hspace{1em}Codebook Learning Rate & $1 \times 10^{-3}$ \\
    \hspace{1em}Commitment Weight ($\beta$) & 1.0 \\
    \hspace{1em}Warmup Steps & 1,000 \\
    \hspace{1em}Architecture (Enc / Dec) & 15 / 15 layers \\
    \hspace{1em}Codebook Size & $[256, 128]$ \\
    \hspace{1em}Number of Codes & 8 per arm \\
    \hspace{1em}Parameters & 150 M \\
    \hspace{1em}Batch Size & 4,096 \\
    \hspace{1em}Training Time & $\sim$190 hours \\
    
    \midrule
    \multicolumn{2}{l}{\textit{\textbf{VD-VAE}}} \\
    \hspace{1em}Total Training Steps & 100,000 \\
    \hspace{1em}VAE Learning Rate & $2 \times 10^{-4}$ \\
    \hspace{1em}Codebook Learning Rate & $1 \times 10^{-4}$ \\
    \hspace{1em}Commitment Weight ($\beta$) & 1.0 \\
    \hspace{1em}Consistency Weight ($\lambda_{\text{con}}$) & 0.1 \\
    \hspace{1em}Regularization Weight ($\lambda_{\text{reg}}$) & 0.5 \\
    \hspace{1em}Warmup Steps & 1,000 \\
    \hspace{1em}Architecture (Enc / Dec) & 12 / 12 layers \\
    \hspace{1em}Task-irrelevant Codes & 2 \\
    \hspace{1em}Parameters & 200 M \\
    \hspace{1em}Batch Size & 256 \\
    \hspace{1em}Training Time & $\sim$380 hours \\
    
    \midrule
    \multicolumn{2}{l}{\textit{\textbf{CLAP-NTP}}} \\
    \hspace{1em}Total Training Steps & 150,000 \\
    \hspace{1em}Peak Learning Rate & $5 \times 10^{-5}$ \\
    \hspace{1em}Min Learning Rate & $5 \times 10^{-6}$ \\
    \hspace{1em}Warmup Steps & 1,000 \\
    \hspace{1em}LR Schedule & Cosine Decay (after 100k) \\
    \hspace{1em}Batch Size & 512 \\
    \hspace{1em}Training Time & $\sim$3,800 hours \\
    
    \midrule
    \multicolumn{2}{l}{\textit{\textbf{\add{Post-training (NTP + CLAP-RF/KM)}}}} \\
    \hspace{1em}\add{Stage 1 Objective} & \add{NTP adaptation} \\
    \hspace{1em}\add{Stage 1 Training Epochs} & \add{3} \\
    \hspace{1em}\add{Stage 1 Learning Rate} & \add{$2 \times 10^{-5}$} \\
    \hspace{1em}\add{Stage 1 Warmup Steps} & \add{1,000} \\
    \hspace{1em}\add{Stage 1 LR Schedule} & \add{No decay} \\
    \hspace{1em}\add{Stage 2 Objective} & \add{RF action expert + KM} \\
    \hspace{1em}\add{Stage 2 Training Epochs} & \add{5} \\
    \hspace{1em}\add{Stage 2 Learning Rate} & \add{$1 \times 10^{-4}$} \\
    \hspace{1em}\add{Stage 2 Warmup Steps} & \add{1,000} \\
    \hspace{1em}\add{Stage 2 LR Schedule} & \add{Cosine decay} \\
    \hspace{1em}\add{Batch Size} & \add{128} \\
    \hspace{1em}\add{Training Time} & \add{Task-dependent} \\
    
    \bottomrule
    \end{tabular}
    }
\end{table}

\bibliography{reference}
\bibliographystyle{IEEEtran}

\vfill

\end{document}